\pdfoutput=1

\documentclass[11pt]{article}

\usepackage{acl}

\usepackage{times}
\usepackage{latexsym}

\usepackage[T1]{fontenc}

\usepackage[utf8]{inputenc}

\usepackage{microtype}

\usepackage{amsfonts}
\usepackage{amsmath}
\usepackage{amssymb}
\usepackage[english]{babel}
\usepackage{color}

\usepackage{graphicx}
\usepackage{float} 
\usepackage{subfigure}
\usepackage{booktabs}
\usepackage{array}
\usepackage{enumitem}
\usepackage{amsthm}

\usepackage{bbm}
\usepackage{multirow}
\usepackage[title]{appendix}

\theoremstyle{definition}

\theoremstyle{remark}

\newcommand\ignore[1]{}

\usepackage{color}
\usepackage{tcolorbox}
\usepackage{CJK}
\usepackage{adjustbox}
\tcbset{width=0.9\textwidth,boxrule=0pt,colback=red,arc=0pt,auto outer arc,left=0pt,right=0pt,boxsep=5pt}

\title{An Empirical Study of Memorization in NLP}

\author{Xiaosen Zheng \\
  Singapore Management University \\
  \texttt{xszheng.2020@phdcs.smu.edu.sg} \\\And
  Jing Jiang \\
  Singapore Management University \\
  \texttt{jingjiang@smu.edu.sg} \\}

\begin{document}
\maketitle
\begin{abstract}
A recent study by \newcite{feldman2020does} proposed a long-tail theory to explain the memorization behavior of deep learning models. However, memorization has not been empirically verified in the context of NLP, a gap addressed by this work. In this paper, we use three different NLP tasks to check if the long-tail theory holds. Our experiments demonstrate that top-ranked memorized training instances are likely atypical, and removing the top-memorized training instances leads to a more serious drop in test accuracy compared with removing training instances randomly. Furthermore, we develop an attribution method to better understand why a training instance is memorized. We empirically show that our memorization attribution method is faithful and share our interesting finding that the top-memorized parts of a training instance tend to be features negatively correlated with the class label.

\end{abstract}

\section{Introduction}
\label{sec:intro}

In recent years, there has been an increasing amount of interest in the machine learning community to understand the \emph{memorization} behaviour of deep neural network models.
Studies have shown that deep learning models often have sufficient capacities to ``memorize'' training examples~\cite{zhang2017understanding, arpit2017closerlook}.
A number of recent studies tried to understand how memorization helps generalization~\cite{chatterjee2018learning, feldman2020does, montanari2020interpolation, khandelwal2019generalization, khandelwal2020nearest}

In NLP, memorization of training examples by deep learning models is also often observed~\cite{li-wisniewski-2021-neural, lewis-etal-2021-question, raunak-etal-2021-curious}, and existing studies usually see memorization as something that hinders generalization.
For example, \newcite{elangovan-etal-2021-memorization} tried to measure the amount of ``data leakage'' in NLP datasets in order to assess a model's ability to memorize vs. its ability to generalize.

However, recently \newcite{feldman2020does} proposed a long-tail theory, which states that memorization is necessary for generalization if the data follows a long-tail distribution.
This theory was later empirically validated by \newcite{feldman2020neural}, but their validation was done in only the computer vision domain.
It is therefore interesting and useful for us to study whether the long-tail theory also holds in NLP;
such validation would help us better understand the utility of memorization in the context of NLP.

The long-tail theory states that if the training data form a long-tail distribution, where there are many small ``sub-populations'' that are atypical instances, and if these small sub-populations are also present in the test data, then memorizing these atypical instances helps the model generalize to the test data.
In order to validate this long-tail theory in the context of NLP, we follow the experiments and analyses on image classification done by \newcite{feldman2020neural}.
Specifically, we aim to answer the following questions in this paper:
(1) On a few typical NLP tasks, are the training instances memorized by deep learning models indeed \emph{atypical} instances?
(2) Does memorizing these training instances lead to \emph{lower generalization error} on the test instances?

In addition, observing that it is not always straightforward to understand why a training instance is being memorized, we study the following novel research question:
(3) Can we provide some explanation about why a training instances is memorized? To be more specific, can we attribute the memorization score of a training instance to its individual tokens such that we can quantify which tokens require the most memorization by the model?

To answer these research questions, we first adopt self-influence~\cite{koh2017understanding} as our memorization scoring function.
Compared with the estimator proposed by~\newcite{feldman2020neural}, our self-influence function is also theoretically motivated but has the advantage that it is easy for us to derive a memorization attribution method for the third research question above.
We present the self-influence function in Section~\ref{subsec:memorization}, and in Section~\ref{subsec:attribution}, we present our novel memorization attribution method.
We conduct experiments on three NLP tasks: sentiment classification, natural language inference~(NLI) and text classification.

Our experiments and analyses demonstrate that \emph{the training instances with the highest memorization scores tend to be atypical}, at least on sentiment classification and NLI.
On all three tasks, we find that removing the top-memorized training instances results in significantly dropped test performance, and the drop is markedly higher compared with removing a random subset of training instances.
We also evaluate our memorization attribution method and find that our method can indeed identify input tokens that require the most memorization.
Finally, we apply our memorization attribution method to sentiment classification and to an image classification dataset, and we share the interesting finding that the highly-memorized input features tend to be those that are negatively correlated with the class labels.
Our code and data are available at \url{https://github.com/xszheng2020/memorization}.

\section{Our Approach}

To validate the long-tail theory in the context NLP, let us first review the main claims of the theory.
First, the long-tail theory hypothesizes that training instances with the same class label has a long-tail distribution, with instances at the tail end being those atypical instances that need to be memorized.
To verify this assumption, we first identify those training instances that are memorized by a trained deep learning model and then check if they are indeed  atypical.
Specifically, we follow \newcite{feldman2020neural} and adopt ``self-influence'' to measure memorization, but we use the influence function proposed by \newcite{koh2017understanding} to define self-influence.
Second, the long tail theory states that memorization of the atypical training instances leads to lower generalization error, because the atypical training instances belong to subpopulations that also have presence in the test data.
To verify this statement, we check whether removing the memorized training instances would lead to more significant performance drop on the test data than removing a random sample of training instances.

It is worth noting that the approach outlined above follows the experiments conducted by \newcite{feldman2020neural} to validate the long tail theory on image classification.

Furthermore, we want to pinpoint which parts of a memorized instance are most critical for memorization.
In other words, since each training instance is assigned a memorization score, can we attribute the memorization score to different parts of the input of this instance?
This presumably can help us better understand which parts of the input need to be memorized the most.
We follow the idea from Integrated Gradients~(IG)~\cite{sundararajan2017axiomatic} and derive a formula to compute memorization attribution.

\subsection{Memorization: Self-Influence}
\label{subsec:memorization}

The high level idea of \newcite{feldman2020does} to define memorization is that memorization measures how the prediction on a training instance $z = (x, y)$ (where $x$ is the observation and $y$ is the label) changes when $z$ is removed from the training data.
This notion is closely related to the influence function defined by \newcite{koh2017understanding}, which measures how much the loss at a test point $z_{\text{test}}$ is influenced by a slight upweighting of a training instance $z$ in the training loss function.
While influence function is generally used to measure the influence of a training instance on a \emph{test} instance, if we use it to measure the influence of a training instance on \emph{itself}, i.e., to measure ``self-influence,'' then this self-influence corresponds to the general notion of memorization defined by \newcite{feldman2020does}.

Adopting the influence function defined by \newcite{koh2017understanding}, we define the memorization score for a training instance $z$ as follows:
\begin{equation}
\begin{aligned}
    \mathcal{M}_{\text{remove}}(z) 
    &\overset{\text{def}}{=} -\frac{d P(y|x; \hat{\theta}_{\epsilon, -z})}{d \epsilon} \bigg|_{\epsilon=0},\\
\end{aligned}
\label{IF.remove}
\end{equation}
where $\hat{\theta}_{\epsilon, -z}$ represents the parameters of the model trained with the instance $z$ down-weighted by $\epsilon$, $P(y|x; \theta)$ is the conditional probability 
using 
$\theta$. 
Thus $\mathcal{M}_{\text{remove}}(z)$ is the amount of change of $P(y|x; \theta)$ when the instance $z$ is down-weighted by a small amount $\epsilon$.

After several steps of derivation (details to be given in Appendix~\ref{appendix:memorizaton_score}), the computation of Eqn~\ref{IF.remove} follows the following formula:
\begin{equation}
\begin{aligned}
    \mathcal{M}_{\text{remove}}(z) 
    &= -\nabla_{\theta}P(y|x; \hat{\theta})^{\top}H^{-1}_{\hat{\theta}}\nabla_{\theta}{L(z, \hat{\theta})},
\end{aligned}
\label{IF.remove.execution}
\end{equation}
where $\hat{\theta}$ is the parameters of the model trained with all instances, $L$ is the loss function (cross entropy in our implementation) and $H_{\hat{\theta}} = \frac{1}{n}\sum^{n}_{i=1}{\nabla^{2}_{\theta}{L(z_i, \hat{\theta})}}$, where $(z_1, z_2, \ldots, z_n)$ are the training instances.

\subsection{Memorization Attribution}
\label{subsec:attribution}

In order to better understand why an instance is memorized, we propose a fine-grained notion of memorization at ``feature'' level instead of instance level, i.e., to attribute the memorization score of an instance to its individual features.
Our proposed memorization attribution method is general and can be applied to any input representation.
For NLP tasks, this means we attribute the memorization score defined above to each token of the input sequence.
For images, this would be to attribute the memorization scores to pixels.

For this memorization attribution, we borrow the idea from Integrated Gradients~(IG)~\cite{sundararajan2017axiomatic}, which is a gradient-based attribution method for understanding which parts of a test instance are more responsible for its prediction.
In particular, the IG method requires an \emph{uninformative} baseline input $x'$ as a reference point.
Similarly, here we also assume a baseline $x'$.
This baseline is supposedly an instance that does not have any influence on any test instance, and in our implementation, we use an sequence of the same length as $x$ but consisting of only the \texttt{[MASK]} tokens.

We first consider the influence of replacing $z = (x, y)$ with the baseline $z' = (x', y)$ (which is similar to perturbation-based influence from~\cite{koh2017understanding}): 

\begin{equation}
\begin{aligned}
    \mathcal{M}_{\text{replace}}(z)
    &\overset{\text{def}}{=} -\frac{d P(y|x; \hat{\theta}_{\epsilon, z', -z})}{d \epsilon} \bigg|_{\epsilon=0}, 
\end{aligned}
\label{IF.replace.1}
\end{equation}
where $\hat{\theta}_{\epsilon, z', -z}$ represents the parameters resulting from moving $\epsilon$ mass from $z$ to $z'$, i.e., adding $z'$ to the training data and giving it a weight of $\epsilon$ in the loss function while reducing the weight of the original $z$ by $\epsilon$.
Thus $\mathcal{M}_{\text{replace}}(z)$ is the amount of change of $P(y|x; \theta)$ when a small amount $\epsilon$ of $z$ is replaced by the uninformative $z'$.

It is worth pointing out that we can regard $\mathcal{M}_{\text{replace}}(z)$ as an alternative way of measuring the amount of memorization of $z$, similar to how perturbation-based influence is an alternative way of measuring influence in~\cite{koh2017understanding}.

With similar derivation steps, the computation of Eqn~\ref{IF.replace.1} is as follows:
\begin{equation}
\begin{aligned}
    \mathcal{M}_{\text{replace}}(z)
    &= -s^{\top} \left(\nabla_{\theta}{L(z, \hat{\theta})} - \nabla_{\theta}{L(z', \hat{\theta})}\right),
\end{aligned}
\label{IF.replace.1.execution}
\end{equation}
where $s=H^{-1}_{\hat{\theta}}\nabla_{\theta}{P(y|x; \hat{\theta})}$. 
(For more details, please refer to Appendix~\ref{appendix:memorizaton_attribution}.)

The advantage of using this alternative measure of memorization is that $\mathcal{M}_{\text{replace}}(z)$ can be decomposed into a linear combination of scores, each corresponding to a single token in the input sequence.
For NLP applications, the input $x$ usually corresponds to an embedding matrix $\mathbf{X} \in \mathbb{R}^{N \times d}$ (where $N$ is the number of tokens and $d$ is the embedding dimensions).
We can show that
\begin{equation}
\begin{aligned}
    \mathcal{M}_{\text{replace}}(z) 
    &= -\sum^{N}_{t=1}{ \sum^{d}_{l=1}{r_{t,l} (\mathbf{X}_{t, l} - \mathbf{X}'_{t, l})}},
\end{aligned}
\label{IF.replace.2}
\end{equation}
where $r=\left[\int^1_{\alpha = 0}{\frac{d g\left(\mathbf{X}' + \alpha (\mathbf{X} - \mathbf{X}') \right)}{d x} d\alpha} \right]s$ and $g(\mathbf{X}) =  \nabla_{\theta}{L((\mathbf{X}, y), \hat{\theta})}$, 
which can be efficiently computed by the hessian-vector product~\cite{pearlmutter1994fast}. 
For more details, please refer to Appendix~\ref{appendix:memorizaton_attribution}. 

The memorization attribution of the $t$-th token is thus given by 
$-\sum^{d}_{l=1}{r_{t,l} \times (\mathbf{X}_{t, l} - \mathbf{X}'_{t, l})}$.

\section{Experiments}
\label{sec:exp}

With the memorization score defined in Eqn~\ref{IF.remove.execution} and the memorization attribution score defined in Eqn~\ref{IF.replace.2}, we now conduct experiments to answer the three research questions raised in Section~\ref{sec:intro}.

\subsection{Experiment Settings}
We conduct our experiments on the following three datasets:

    \noindent \textbf{SST-2}~\cite{socher-etal-2013-recursive}: This is a dataset for sentence-level binary (positive vs. negative) sentiment classification.
    It consists of 6,920 training instances, 872 development instances and 1,821 test instances.

    \noindent \textbf{SNLI}~\cite{maccartney-manning-2008-modeling}: This is a dataset for natural language inference, which aims to predict the entailment relation (contradiction, neutral or entailment) between a premise and a hypothesis. 
    We combine the \emph{contradiction} and \emph{neutral} classes into a single \emph{non-entailment} class, and randomly sample 10k training instances, 6,658 development instances and 6,736 test instances.
    
    \noindent \textbf{Yahoo! Answers}~\cite{zhang2015character}: This is a collection of question-answer pairs categorized into 10 topic-based classes. 
    We randomly sample 10k training instances, 10k development instances and 10k test examples.

In addition, 
we also use \textbf{CIFAR-10}~\cite{krizhevsky2009learning}, which is a dataset for 10-class image classification. 
We randomly sample 10k training instances, 5k development instances and 10k test instances.
For some tasks, we down-sample the training set because influence function is known to be expensive to compute.

For all NLP tasks, we adopt the pre-trained Distill-BERT model~\cite{sanh2019distilbert} that consists of 6 transformer layers, where each layer consists of 12 attention heads. 
We use the final hidden state of the \texttt{[CLS]} token for classification.\footnote{Following \newcite{han-etal-2020-explaining, guo-etal-2021-fastif}, we ``freeze" the word embedding layer and the first 4 transformer layers, only fine-tuning the last 2 transformer layers and the final projection layer because of the computation limits.}
For CIFAR-10, we extract visual grid features using a pre-trained ResNet50~\cite{he2016deep} first and then train a MLP classifier on top of that.

We use the SGD optimizer,
setting the learning rate, momentum and batch size to 0.01, 0.9 and 32, respectively. 
We tune other hyper-parameters on the development set manually.

Although influence function is model-dependent and therefore models trained with different random seeds may produce different memorization scores for the same training instance, we found that in practice, ranking training instances based on memorization scores obtained from models trained by different random seeds produces similar rankings across different models.
Thus, we only consider a single model checkpoint for computing our self-influence based memorization scores in the following experiments. (See Appendix~\ref{appendix:checkpoints} for the exact description.)
For memorization attribution, the number of Riemann Sum steps is set to be 50. 

\subsection{Checking Memorized Instances}
\begin{table}[htbp]
\centering

\begin{small}
\begin{tabular}{lll}
\toprule
\textbf{Group} & \textbf{Negative} & \textbf{Positive} \\
\midrule
Top-10\% & 35.80 & 74.00 \\
\midrule
All & 23.24 & 86.39 \\
\midrule
Bottom-10\% & 14.92 & 94.52 \\
\bottomrule
\end{tabular}
\end{small}

\caption{The average percentage of \emph{positive phrases} over (1) the top-10\% memorized positive/negative instances, (2) all positive/negative instances, and (3) the bottom-10\% memorized positive/negative instances. 
}
\label{table.atypical}
\end{table}

\begin{table*}
\centering
\resizebox{\textwidth}{!}{
\begin{small}
\begin{tabular}{llll}
\toprule
\multicolumn{2}{c}{\textbf{Negative}} & \multicolumn{2}{c}{\textbf{Positive}} \\
\cmidrule(lr){1-2} \cmidrule(lr){3-4}
\textbf{Content} &\textbf{Mem} & \textbf{Content} &\textbf{Mem}\\
\midrule
\begin{CJK*}{UTF8}{gbsn}
{\setlength{\fboxsep}{0pt}\colorbox{white!0}{\parbox{0.65\textwidth}{
\colorbox{red!0}{\strut Starts} \colorbox{red!0}{\strut out} \colorbox{red!0}{\strut with} \colorbox{red!0}{\strut tremendous} \colorbox{red!0}{\strut promise,} \colorbox{red!0}{\strut introducing} \colorbox{red!0}{\strut an} \colorbox{red!0}{\strut intriguing} \colorbox{red!0}{\strut and} \colorbox{red!0}{\strut alluring} \colorbox{red!0}{\strut premise,} \colorbox{red!0}{\strut only} \colorbox{red!0}{\strut to} \colorbox{red!0}{\strut fall} \colorbox{red!0}{\strut prey} \colorbox{red!0}{\strut to} \colorbox{red!0}{\strut a} \colorbox{red!0}{\strut boatload} \colorbox{red!0}{\strut of} \colorbox{red!0}{\strut screenwriting} \colorbox{red!0}{\strut cliches} \colorbox{red!0}{\strut that} \colorbox{red!0}{\strut sink} \colorbox{red!0}{\strut it} \colorbox{red!0}{\strut faster} \colorbox{red!0}{\strut than} \colorbox{red!0}{\strut a} \colorbox{red!0}{\strut leaky} \colorbox{red!0}{\strut freighter} 
}}}
\end{CJK*} 
& 14.83 & 
\begin{CJK*}{UTF8}{gbsn}
{\setlength{\fboxsep}{0pt}\colorbox{white!0}{\parbox{0.65\textwidth}{
\colorbox{red!0}{\strut The} \colorbox{red!0}{\strut director,} \colorbox{red!0}{\strut Mark} \colorbox{red!0}{\strut Pellington,} \colorbox{red!0}{\strut does} \colorbox{red!0}{\strut a} \colorbox{red!0}{\strut terrific} \colorbox{red!0}{\strut job} \colorbox{red!0}{\strut conjuring} \colorbox{red!0}{\strut up} \colorbox{red!0}{\strut a} \colorbox{red!0}{\strut sinister,} \colorbox{red!0}{\strut menacing} \colorbox{red!0}{\strut atmosphere} \colorbox{red!0}{\strut though} \colorbox{red!0}{\strut unfortunately} \colorbox{red!0}{\strut all} \colorbox{red!0}{\strut the} \colorbox{red!0}{\strut story} \colorbox{red!0}{\strut gives} \colorbox{red!0}{\strut us} \colorbox{red!0}{\strut is} \colorbox{red!0}{\strut flashing} \colorbox{red!0}{\strut red} \colorbox{red!0}{\strut lights,} \colorbox{red!0}{\strut a} \colorbox{red!0}{\strut rattling} \colorbox{red!0}{\strut noise,} \colorbox{red!0}{\strut and} \colorbox{red!0}{\strut a} \colorbox{red!0}{\strut bump} \colorbox{red!0}{\strut on} \colorbox{red!0}{\strut the} \colorbox{red!0}{\strut head} 
}}}\end{CJK*} 
& 14.28\\ 
\begin{CJK*}{UTF8}{gbsn}
{\setlength{\fboxsep}{0pt}\colorbox{white!0}{\parbox{0.65\textwidth}{
\colorbox{red!0}{\strut Mr.} \colorbox{red!0}{\strut Wollter} \colorbox{red!0}{\strut and} \colorbox{red!0}{\strut Ms.} \colorbox{red!0}{\strut Seldhal} \colorbox{red!0}{\strut give} \colorbox{red!0}{\strut strong} \colorbox{red!0}{\strut and} \colorbox{red!0}{\strut convincing} \colorbox{red!0}{\strut performances,} \colorbox{red!0}{\strut but} \colorbox{red!0}{\strut neither} \colorbox{red!0}{\strut reaches} \colorbox{red!0}{\strut into} \colorbox{red!0}{\strut the} \colorbox{red!0}{\strut deepest} \colorbox{red!0}{\strut recesses} \colorbox{red!0}{\strut of} \colorbox{red!0}{\strut the} \colorbox{red!0}{\strut character} \colorbox{red!0}{\strut to} \colorbox{red!0}{\strut unearth} \colorbox{red!0}{\strut the} \colorbox{red!0}{\strut quaking} \colorbox{red!0}{\strut essence} \colorbox{red!0}{\strut of} \colorbox{red!0}{\strut passion,} \colorbox{red!0}{\strut grief} \colorbox{red!0}{\strut and} \colorbox{red!0}{\strut fear} 
}}}
\end{CJK*} 
& 13.65 &
\begin{CJK*}{UTF8}{gbsn}
{\setlength{\fboxsep}{0pt}\colorbox{white!0}{\parbox{0.65\textwidth}{
\colorbox{red!0}{\strut This} \colorbox{red!0}{\strut is} \colorbox{red!0}{\strut a} \colorbox{red!0}{\strut fascinating} \colorbox{red!0}{\strut film} \colorbox{red!0}{\strut because} \colorbox{red!0}{\strut there} \colorbox{red!0}{\strut is} \colorbox{red!0}{\strut no} \colorbox{red!0}{\strut clear-cut} \colorbox{red!0}{\strut hero} \colorbox{red!0}{\strut and} \colorbox{red!0}{\strut no} \colorbox{red!0}{\strut all-out} \colorbox{red!0}{\strut villain} 
}}}\end{CJK*}
& 14.18 \\ 
\begin{CJK*}{UTF8}{gbsn}
{\setlength{\fboxsep}{0pt}\colorbox{white!0}{\parbox{0.65\textwidth}{
\colorbox{red!0}{\strut This} \colorbox{red!0}{\strut is} \colorbox{red!0}{\strut a} \colorbox{red!0}{\strut monumental} \colorbox{red!0}{\strut achievement} \colorbox{red!0}{\strut in} \colorbox{red!0}{\strut practically} \colorbox{red!0}{\strut every} \colorbox{red!0}{\strut facet} \colorbox{red!0}{\strut of} \colorbox{red!0}{\strut inept} \colorbox{red!0}{\strut filmmaking:} \colorbox{red!0}{\strut joyless,} \colorbox{red!0}{\strut idiotic,} \colorbox{red!0}{\strut annoying,} \colorbox{red!0}{\strut heavy-handed} \colorbox{red!0}{\strut visually} \colorbox{red!0}{\strut atrocious,} \colorbox{red!0}{\strut and} \colorbox{red!0}{\strut often} \colorbox{red!0}{\strut downright} \colorbox{red!0}{\strut creepy} 
}}}
\end{CJK*} 
& 11.01 &
\begin{CJK*}{UTF8}{gbsn}
{\setlength{\fboxsep}{0pt}\colorbox{white!0}{\parbox{0.65\textwidth}{
\colorbox{red!0}{\strut The} \colorbox{red!0}{\strut film} \colorbox{red!0}{\strut is} \colorbox{red!0}{\strut reasonably} \colorbox{red!0}{\strut entertaining,} \colorbox{red!0}{\strut though} \colorbox{red!0}{\strut it} \colorbox{red!0}{\strut begins} \colorbox{red!0}{\strut to} \colorbox{red!0}{\strut drag} \colorbox{red!0}{\strut two-thirds} \colorbox{red!0}{\strut through,} \colorbox{red!0}{\strut when} \colorbox{red!0}{\strut the} \colorbox{red!0}{\strut melodramatic} \colorbox{red!0}{\strut aspects} \colorbox{red!0}{\strut start} \colorbox{red!0}{\strut to} \colorbox{red!0}{\strut overtake} \colorbox{red!0}{\strut the} \colorbox{red!0}{\strut comedy} 
}}}\end{CJK*}
& 11.04 \\ 
\midrule
\begin{CJK*}{UTF8}{gbsn}
{\setlength{\fboxsep}{0pt}\colorbox{white!0}{\parbox{0.65\textwidth}{
\colorbox{red!0}{\strut Sadly,} \colorbox{red!0}{\strut Full} \colorbox{red!0}{\strut Frontal} \colorbox{red!0}{\strut plays} \colorbox{red!0}{\strut like} \colorbox{red!0}{\strut the} \colorbox{red!0}{\strut work} \colorbox{red!0}{\strut of} \colorbox{red!0}{\strut a} \colorbox{red!0}{\strut dilettante} 
}}}\end{CJK*} 
& 0.00 &
\begin{CJK*}{UTF8}{gbsn}
{\setlength{\fboxsep}{0pt}\colorbox{white!0}{\parbox{0.65\textwidth}{
\colorbox{red!0}{\strut The} \colorbox{red!0}{\strut large-format} \colorbox{red!0}{\strut film} \colorbox{red!0}{\strut is} \colorbox{red!0}{\strut well} \colorbox{red!0}{\strut suited} \colorbox{red!0}{\strut to} \colorbox{red!0}{\strut capture} \colorbox{red!0}{\strut these} \colorbox{red!0}{\strut musicians} \colorbox{red!0}{\strut in} \colorbox{red!0}{\strut full} \colorbox{red!0}{\strut regalia} \colorbox{red!0}{\strut and} \colorbox{red!0}{\strut the} \colorbox{red!0}{\strut incredible} \colorbox{red!0}{\strut IMAX} \colorbox{red!0}{\strut sound} \colorbox{red!0}{\strut system} \colorbox{red!0}{\strut lets} \colorbox{red!0}{\strut you} \colorbox{red!0}{\strut feel} \colorbox{red!0}{\strut the} \colorbox{red!0}{\strut beat} \colorbox{red!0}{\strut down} \colorbox{red!0}{\strut to} \colorbox{red!0}{\strut your} \colorbox{red!0}{\strut toes} 
}}}\end{CJK*} & 0.00 \\ 
\begin{CJK*}{UTF8}{gbsn}
{\setlength{\fboxsep}{0pt}\colorbox{white!0}{\parbox{0.65\textwidth}{
\colorbox{red!0}{\strut A} \colorbox{red!0}{\strut mess}
}}}\end{CJK*} 
& 0.00 &
\begin{CJK*}{UTF8}{gbsn}
{\setlength{\fboxsep}{0pt}\colorbox{white!0}{\parbox{0.65\textwidth}{
\colorbox{red!0}{\strut P.T.} \colorbox{red!0}{\strut Anderson} \colorbox{red!0}{\strut understands} \colorbox{red!0}{\strut the} \colorbox{red!0}{\strut grandness} \colorbox{red!0}{\strut of} \colorbox{red!0}{\strut romance} \colorbox{red!0}{\strut and} \colorbox{red!0}{\strut how} \colorbox{red!0}{\strut love} \colorbox{red!0}{\strut is} \colorbox{red!0}{\strut the} \colorbox{red!0}{\strut great} \colorbox{red!0}{\strut equalizer} \colorbox{red!0}{\strut that} \colorbox{red!0}{\strut can} \colorbox{red!0}{\strut calm} \colorbox{red!0}{\strut us} \colorbox{red!0}{\strut of} \colorbox{red!0}{\strut our} \colorbox{red!0}{\strut daily} \colorbox{red!0}{\strut ills} \colorbox{red!0}{\strut and} \colorbox{red!0}{\strut bring} \colorbox{red!0}{\strut out} \colorbox{red!0}{\strut joys} \colorbox{red!0}{\strut in} \colorbox{red!0}{\strut our} \colorbox{red!0}{\strut lives} \colorbox{red!0}{\strut that} \colorbox{red!0}{\strut we} \colorbox{red!0}{\strut never} \colorbox{red!0}{\strut knew} \colorbox{red!0}{\strut were} \colorbox{red!0}{\strut possible} 
}}}\end{CJK*}
& 0.00 \\ 
\begin{CJK*}{UTF8}{gbsn}
{\setlength{\fboxsep}{0pt}\colorbox{white!0}{\parbox{0.65\textwidth}{
\colorbox{red!0}{\strut The} \colorbox{red!0}{\strut images} \colorbox{red!0}{\strut lack} \colorbox{red!0}{\strut contrast,} \colorbox{red!0}{\strut are} \colorbox{red!0}{\strut murky} \colorbox{red!0}{\strut and} \colorbox{red!0}{\strut are} \colorbox{red!0}{\strut frequently} \colorbox{red!0}{\strut too} \colorbox{red!0}{\strut dark} \colorbox{red!0}{\strut to} \colorbox{red!0}{\strut be} \colorbox{red!0}{\strut decipherable}  
}}}\end{CJK*}
& 0.00 &
\begin{CJK*}{UTF8}{gbsn}
{\setlength{\fboxsep}{0pt}\colorbox{white!0}{\parbox{0.65\textwidth}{
\colorbox{red!0}{\strut together} \colorbox{red!0}{\strut writer-director} \colorbox{red!0}{\strut Danny} \colorbox{red!0}{\strut Verete's} \colorbox{red!0}{\strut three} \colorbox{red!0}{\strut tales} \colorbox{red!0}{\strut comprise} \colorbox{red!0}{\strut a} \colorbox{red!0}{\strut powerful} \colorbox{red!0}{\strut and} \colorbox{red!0}{\strut reasonably} \colorbox{red!0}{\strut fulfilling} \colorbox{red!0}{\strut gestalt} 
}}}\end{CJK*} 
& 0.00 \\
\bottomrule
\end{tabular}
\end{small}
}
\caption{Top-3 and Bottom-3 memorized training examples from the SST-2 task. Note that there are many examples having zero memorization score, we randomly sample 3 out of them.
}
\label{table.vis.sst}
\end{table*}

\begin{table*}
\centering
\resizebox{\textwidth}{!}{
\begin{small}
\begin{tabular}{llll}
\toprule
\multicolumn{2}{c}{\textbf{Non-Entail}} & \multicolumn{2}{c}{\textbf{Entail}} \\
\cmidrule(lr){1-2} \cmidrule(lr){3-4}
\textbf{Content} &\textbf{Mem} & \textbf{Content} &\textbf{Mem}\\
\midrule
\begin{CJK*}{UTF8}{gbsn}
{\setlength{\fboxsep}{0pt}\colorbox{white!0}{\parbox{0.65\textwidth}{
\textbf{P:} \colorbox{red!0}{\strut A} \colorbox{red!0}{\strut man} \colorbox{red!0}{\strut in} \colorbox{red!0}{\strut a} \colorbox{red!0}{\strut bright} \colorbox{red!0}{\strut pastel} \colorbox{red!0}{\strut blue} \colorbox{red!0}{\strut overcoat} \colorbox{red!0}{\strut plays} \colorbox{red!0}{\strut a} \colorbox{red!0}{\strut unique} \colorbox{red!0}{\strut instrument} \colorbox{red!0}{\strut by} \colorbox{red!0}{\strut the} \colorbox{red!0}{\strut corner} \colorbox{red!0}{\strut of} \colorbox{red!0}{\strut a} \colorbox{red!0}{\strut building} \colorbox{red!0}{\strut with} \colorbox{red!0}{\strut a} \colorbox{red!0}{\strut sign} \colorbox{red!0}{\strut propped} \colorbox{red!0}{\strut against} \colorbox{red!0}{\strut a} \colorbox{red!0}{\strut bag} \colorbox{red!0}{\strut in} \colorbox{red!0}{\strut front} \colorbox{red!0}{\strut of} \colorbox{red!0}{\strut him}
\\ \textbf{H:} 
\colorbox{red!0}{\strut A} \colorbox{red!0}{\strut man} \colorbox{red!0}{\strut plays} \colorbox{red!0}{\strut a} \colorbox{red!0}{\strut guitar} \colorbox{red!0}{\strut outside}
}}}
\end{CJK*}
& 18.85 &
\begin{CJK*}{UTF8}{gbsn}
{\setlength{\fboxsep}{0pt}\colorbox{white!0}{\parbox{0.65\textwidth}{
\textbf{P:} 
\colorbox{red!0}{\strut An} \colorbox{red!0}{\strut older} \colorbox{red!0}{\strut man} \colorbox{red!0}{\strut in} \colorbox{red!0}{\strut a} \colorbox{red!0}{\strut white} \colorbox{red!0}{\strut shirt} \colorbox{red!0}{\strut is} \colorbox{red!0}{\strut playing} \colorbox{red!0}{\strut a} \colorbox{red!0}{\strut keyboard}
\\ \textbf{H:} 
\colorbox{red!0}{\strut A} \colorbox{red!0}{\strut man} \colorbox{red!0}{\strut is} \colorbox{red!0}{\strut playing} \colorbox{red!0}{\strut the} \colorbox{red!0}{\strut piano} 
}}}
\end{CJK*}
& 23.24 \\ 
\begin{CJK*}{UTF8}{gbsn}
{\setlength{\fboxsep}{0pt}\colorbox{white!0}{\parbox{0.65\textwidth}{
\textbf{P:} 
\colorbox{red!0}{\strut A} \colorbox{red!0}{\strut young} \colorbox{red!0}{\strut boy} \colorbox{red!0}{\strut in} \colorbox{red!0}{\strut a} \colorbox{red!0}{\strut yellow} \colorbox{red!0}{\strut rash} \colorbox{red!0}{\strut guard} \colorbox{red!0}{\strut is} \colorbox{red!0}{\strut walking} \colorbox{red!0}{\strut on} \colorbox{red!0}{\strut the} \colorbox{red!0}{\strut shore} \colorbox{red!0}{\strut carrying} \colorbox{red!0}{\strut a} \colorbox{red!0}{\strut surfboard}
\\ \textbf{H:} 
\colorbox{red!0}{\strut A} \colorbox{red!0}{\strut boy} \colorbox{red!0}{\strut is} \colorbox{red!0}{\strut walking} \colorbox{red!0}{\strut on} \colorbox{red!0}{\strut the} \colorbox{red!0}{\strut boardwalk} 
}}}
\end{CJK*}
& 17.51 & 
\begin{CJK*}{UTF8}{gbsn}
{\setlength{\fboxsep}{0pt}\colorbox{white!0}{\parbox{0.65\textwidth}{
\textbf{P:} 
\colorbox{red!0}{\strut A} \colorbox{red!0}{\strut woman} \colorbox{red!0}{\strut in} \colorbox{red!0}{\strut a} \colorbox{red!0}{\strut white} \colorbox{red!0}{\strut and} \colorbox{red!0}{\strut light} \colorbox{red!0}{\strut green} \colorbox{red!0}{\strut jacket} \colorbox{red!0}{\strut and} \colorbox{red!0}{\strut another} \colorbox{red!0}{\strut woman} \colorbox{red!0}{\strut in} \colorbox{red!0}{\strut a} \colorbox{red!0}{\strut purple} \colorbox{red!0}{\strut shirt} \colorbox{red!0}{\strut ,} \colorbox{red!0}{\strut both} \colorbox{red!0}{\strut wearing} \colorbox{red!0}{\strut hats} \colorbox{red!0}{\strut ,} \colorbox{red!0}{\strut sit} \colorbox{red!0}{\strut at} \colorbox{red!0}{\strut a} \colorbox{red!0}{\strut table} \colorbox{red!0}{\strut watching} \colorbox{red!0}{\strut a} \colorbox{red!0}{\strut cooking} \colorbox{red!0}{\strut fire}
\\ \textbf{H:} 
\colorbox{red!0}{\strut A} \colorbox{red!0}{\strut woman} \colorbox{red!0}{\strut in} \colorbox{red!0}{\strut a} \colorbox{red!0}{\strut white} \colorbox{red!0}{\strut and} \colorbox{red!0}{\strut light} \colorbox{red!0}{\strut green} \colorbox{red!0}{\strut jacket} 
}}}
\end{CJK*}
& 18.94 \\ 
\begin{CJK*}{UTF8}{gbsn}
{\setlength{\fboxsep}{0pt}\colorbox{white!0}{\parbox{0.65\textwidth}{
\textbf{P:} 
\colorbox{red!0}{\strut Someone} \colorbox{red!0}{\strut wearing} \colorbox{red!0}{\strut a} \colorbox{red!0}{\strut blue} \colorbox{red!0}{\strut shirt} \colorbox{red!0}{\strut is} \colorbox{red!0}{\strut riding} \colorbox{red!0}{\strut a} \colorbox{red!0}{\strut bike} \colorbox{red!0}{\strut with} \colorbox{red!0}{\strut a} \colorbox{red!0}{\strut child} \colorbox{red!0}{\strut '} \colorbox{red!0}{\strut s} \colorbox{red!0}{\strut seat} \colorbox{red!0}{\strut on} \colorbox{red!0}{\strut the} \colorbox{red!0}{\strut front} \colorbox{red!0}{\strut of} \colorbox{red!0}{\strut it}
\\ \textbf{H:} 
\colorbox{red!0}{\strut A} \colorbox{red!0}{\strut person} \colorbox{red!0}{\strut is} \colorbox{red!0}{\strut riding} \colorbox{red!0}{\strut a} \colorbox{red!0}{\strut bike} \colorbox{red!0}{\strut on} \colorbox{red!0}{\strut the} \colorbox{red!0}{\strut street}
}}}
\end{CJK*}
& 15.52 &
\begin{CJK*}{UTF8}{gbsn}
{\setlength{\fboxsep}{0pt}\colorbox{white!0}{\parbox{0.65\textwidth}{
\textbf{P:} 
\colorbox{red!0}{\strut A} \colorbox{red!0}{\strut man} \colorbox{red!0}{\strut sits} \colorbox{red!0}{\strut on} \colorbox{red!0}{\strut a} \colorbox{red!0}{\strut folding} \colorbox{red!0}{\strut chair} \colorbox{red!0}{\strut outside} \colorbox{red!0}{\strut while} \colorbox{red!0}{\strut listening} \colorbox{red!0}{\strut to} \colorbox{red!0}{\strut music} \colorbox{red!0}{\strut on} \colorbox{red!0}{\strut his} \colorbox{red!0}{\strut iPod}
\\ \textbf{H:} 
\colorbox{red!0}{\strut There} \colorbox{red!0}{\strut is} \colorbox{red!0}{\strut a} \colorbox{red!0}{\strut man} \colorbox{red!0}{\strut on} \colorbox{red!0}{\strut a} \colorbox{red!0}{\strut chair} \colorbox{red!0}{\strut listening} \colorbox{red!0}{\strut to} \colorbox{red!0}{\strut music} \colorbox{red!0}{\strut on} \colorbox{red!0}{\strut an} \colorbox{red!0}{\strut mp3} \colorbox{red!0}{\strut player}
}}}
\end{CJK*}
& 18.89 \\
\midrule
\begin{CJK*}{UTF8}{gbsn}
{\setlength{\fboxsep}{0pt}\colorbox{white!0}{\parbox{0.65\textwidth}{
\textbf{P:} 
\colorbox{red!0}{\strut A} \colorbox{red!0}{\strut brunette} \colorbox{red!0}{\strut woman} \colorbox{red!0}{\strut does} \colorbox{red!0}{\strut a} \colorbox{red!0}{\strut wheelie} \colorbox{red!0}{\strut on} \colorbox{red!0}{\strut a} \colorbox{red!0}{\strut white} \colorbox{red!0}{\strut bicycle} \colorbox{red!0}{\strut with} \colorbox{red!0}{\strut purple} \colorbox{red!0}{\strut tires}
\\ \textbf{H:} 
\colorbox{red!0}{\strut A} \colorbox{red!0}{\strut woman} \colorbox{red!0}{\strut rides} \colorbox{red!0}{\strut her} \colorbox{red!0}{\strut motorcycle} \colorbox{red!0}{\strut to} \colorbox{red!0}{\strut town}
}}}
\end{CJK*}
& 0.00 & 
\begin{CJK*}{UTF8}{gbsn}
{\setlength{\fboxsep}{0pt}\colorbox{white!0}{\parbox{0.65\textwidth}{
\textbf{P:} 
\colorbox{red!0}{\strut A} \colorbox{red!0}{\strut married} \colorbox{red!0}{\strut man} \colorbox{red!0}{\strut is} \colorbox{red!0}{\strut taking} \colorbox{red!0}{\strut pictures} \colorbox{red!0}{\strut while} \colorbox{red!0}{\strut standing} \colorbox{red!0}{\strut in} \colorbox{red!0}{\strut a} \colorbox{red!0}{\strut crowd} \colorbox{red!0}{\strut of} \colorbox{red!0}{\strut people}
\\ \textbf{H:} 
\colorbox{red!0}{\strut There} \colorbox{red!0}{\strut are} \colorbox{red!0}{\strut people} \colorbox{red!0}{\strut in} \colorbox{red!0}{\strut a} \colorbox{red!0}{\strut crowd}
}}}
\end{CJK*}
& 0.00 \\ 
\begin{CJK*}{UTF8}{gbsn}
{\setlength{\fboxsep}{0pt}\colorbox{white!0}{\parbox{0.65\textwidth}{
\textbf{P:} 
\colorbox{red!0}{\strut A} \colorbox{red!0}{\strut baseball} \colorbox{red!0}{\strut player} \colorbox{red!0}{\strut hitting} \colorbox{red!0}{\strut a} \colorbox{red!0}{\strut home} \colorbox{red!0}{\strut run}
\\ \textbf{H:} 
\colorbox{red!0}{\strut The} \colorbox{red!0}{\strut cat} \colorbox{red!0}{\strut eats} \colorbox{red!0}{\strut sheep} 
}}}
\end{CJK*}
& 0.00 &
\begin{CJK*}{UTF8}{gbsn}
{\setlength{\fboxsep}{0pt}\colorbox{white!0}{\parbox{0.65\textwidth}{
\textbf{P:} 
\colorbox{red!0}{\strut A} \colorbox{red!0}{\strut man} \colorbox{red!0}{\strut recreates} \colorbox{red!0}{\strut a} \colorbox{red!0}{\strut joust} \colorbox{red!0}{\strut from} \colorbox{red!0}{\strut mid} \colorbox{red!0}{\strut -} \colorbox{red!0}{\strut evil} \colorbox{red!0}{\strut times}
\\ \textbf{H:} 
\colorbox{red!0}{\strut A} \colorbox{red!0}{\strut person} \colorbox{red!0}{\strut created} \colorbox{red!0}{\strut something}
}}}
\end{CJK*}
& 0.00 \\ 
\begin{CJK*}{UTF8}{gbsn}
{\setlength{\fboxsep}{0pt}\colorbox{white!0}{\parbox{0.65\textwidth}{
\textbf{P:} 
\colorbox{red!0}{\strut A} \colorbox{red!0}{\strut child} \colorbox{red!0}{\strut in} \colorbox{red!0}{\strut a} \colorbox{red!0}{\strut vest} \colorbox{red!0}{\strut and} \colorbox{red!0}{\strut hat} \colorbox{red!0}{\strut is} \colorbox{red!0}{\strut posing} \colorbox{red!0}{\strut for} \colorbox{red!0}{\strut a} \colorbox{red!0}{\strut picture}
\\ \textbf{H:} 
\colorbox{red!0}{\strut A} \colorbox{red!0}{\strut child} \colorbox{red!0}{\strut is} \colorbox{red!0}{\strut eating} \colorbox{red!0}{\strut his} \colorbox{red!0}{\strut lunch}
}}}
\end{CJK*}
& 0.00 & 
\begin{CJK*}{UTF8}{gbsn}
{\setlength{\fboxsep}{0pt}\colorbox{white!0}{\parbox{0.65\textwidth}{
\textbf{P:} 
\colorbox{red!0}{\strut A} \colorbox{red!0}{\strut boy} \colorbox{red!0}{\strut is} \colorbox{red!0}{\strut wearing} \colorbox{red!0}{\strut a} \colorbox{red!0}{\strut red} \colorbox{red!0}{\strut towel} \colorbox{red!0}{\strut standing} \colorbox{red!0}{\strut on} \colorbox{red!0}{\strut the} \colorbox{red!0}{\strut beach}
\\ \textbf{H:} 
\colorbox{red!0}{\strut A} \colorbox{red!0}{\strut person} \colorbox{red!0}{\strut is} \colorbox{red!0}{\strut at} \colorbox{red!0}{\strut the} \colorbox{red!0}{\strut beach}
}}}
\end{CJK*}
& 0.00 \\ 
\bottomrule
\end{tabular}
\end{small}
}
\caption{Top-3 and Bottom-3 memorized training examples from the SNLI task. Note that there are many examples having zero memorization score, we randomly sample 3 out of them.
}
\label{table.vis.snli}
\end{table*}

In the first set of experiments, we use our self-influence-based memorization scoring function as defined in Eqn.~\ref{IF.remove} to rank the training instances.

Our goal is to check if the top-memorized instances are indeed atypical instances. 
However, it is difficult to measure the typicality of instances. 
We note that in the prior work~\cite{feldman2020neural} where the authors tried to validate the long-tail theory on computer vision datasets, there was not any quantitative experiment, and the authors relied only on qualitative analysis (i.e., manual inspection of the top-ranked instances) to show that memorized instances tend to be atypical.
In our experiments, we perform two kinds of checking:
(1) First, we adopt qualitative evaluation as \newcite{feldman2020neural} did on both SST-2 and SNLI. 
For Yahoo! Answers, however, because each instance contains a long document, it is not easy for humans to judge whether or not an instance is atypical.
(2) Second, we define quantitative measures of typicality on sentiment analysis because annotations are available on this dataset and these annotations allow us to define some form of typicality.

\subsubsection*{SST-2}
For SST-2, we judge whether or not the top-ranked memorized instances are atypical in two ways:
(1) The first is based on a heuristic metric. 
We check the percentage of positive phrases in an instance, where phrase-level sentiment polarity labels are from the annotations provided by SST-2.
Intuitively, a typical positive sentence should have a relatively high percentage of positive phrases and a typical negative sentence should have a relatively low percentage of positive phrases.
We collect such statistics from SST-2 based on the phrase-level annotations and found that this is to a large extent true.
For example, more than 75\% of positive sentences have at least 78.31\% of positive phrases and more than 75\% of negative sentences have at most 35.73\% of positive phrases.
(See Appendix~\ref{appendix:atypical} for details.)
Therefore, by checking the percentage of positive phrases inside a positive or negative instance, we can in a way judge whether that instance is typical or atypical.
When calculating the percentage of positive phrases inside a sentence, we apply Laplace smoothing.
(2) We also manually inspect the top-ranked and bottom-ranked training instances based on the memorization scores and use our human knowledge to judge whether the top-ranked ones are atypical while the bottom-ranked ones are typical.

Table~\ref{table.atypical} shows the average percentage of positive phrases in the top-10\% of the memorized positive (or negative) training instances and the bottom-10\% of the memorized positive (or negative) training instances.
As a reference point, we also show the average percentage over all positive (or negative) training instances.
We can see that the top-10\% memorized instances indeed are atypical.
Specifically, those negative sentences with high memorization scores have a high percentage of positive phrases on average~(35.80\%), clearly higher than the average percentage of positive phrases of all negative instances~(23.24\%).
This makes the top-memorized negative instances very different from typical negative instances.
On the other hand, the bottom-10\% negative instances (i.e., those instances that are not memorized) have clearly much lower percentage of positive phrases~(14.92\%), which is what we expect for typical negative instances.
Similar observations can be made with the positive training instances.
Overall, the results in Table~\ref{table.atypical} suggest that indeed the top-memorized training instances in SST-2 are atypical.

Next, we manually inspect the top-ranked and bottom-ranked training instances of SST-2 in Table~\ref{table.vis.sst}.
We can see that the top-ranked memorized instances tend to express their overall opinions in an indirect way.
These sentences often contain a contrast between positive and negative opinions.
We therefore believe that they are atypical for sentiment classification.
On the other hand, the bottom-ranked instances, i.e., those with 0 memorization scores, tend to directly expression their opinions with strong opinion phrases, and we believe these represent common instances.

\subsubsection*{SNLI}
For the task of natural language inference, it is hard to come up with a heuristic metric like the one used for sentiment classification.
We therefore focus on manual inspection of the top-ranked and bottom-ranked training instances.
In Table~\ref{table.vis.snli} we show the top-3 and bottom-3 memorized training instances from SNLI.
We can see from the table that in the top-ranked memorized non-entailment instances, the hypothesis tends to be much shorter than the premise and there tends to be no obvious contradiction.
In contrast, the bottom-ranked non-entailment instances tend to be contradictions where there are obvious contradictory words/phrases in the premise and the hypothesis, such as ``bicycle'' vs. ``motorcycle,'' ``player'' vs. ``cat'' and ``posing for a picture'' vs. ``eating his lunch.'' 
We hypothesize that the top-ranked non-entailment instances are atypical because they do not have obvious signals of non-entailment such as the contradictory word pairs we see in the bottom-ranked non-entailment instances.
For entailment cases, we find that the top-ranked instances often contain word pairs that are synonyms but are rare in the training data.
For example, we find that the word pair ``keyboard'' and ``piano'' appears only two times in the training data, which implies that this instance 
is an atypical example.
Similarly, we find that the word/phrase pair ``iPod'' and ``mp3 player'' appear only once in the training data.
On the other hand, the bottom-ranked entailment instances tend to be those where the hypothesis contains less information than the premise, which may be a common type of entailment instances.

\subsection{Marginal Utility of Memorized Instances}
\begin{figure*}[t]
\centering
\subfigure[SST-2]{
\label{Fig.Marginal.1}
\includegraphics[width=0.235\textwidth]{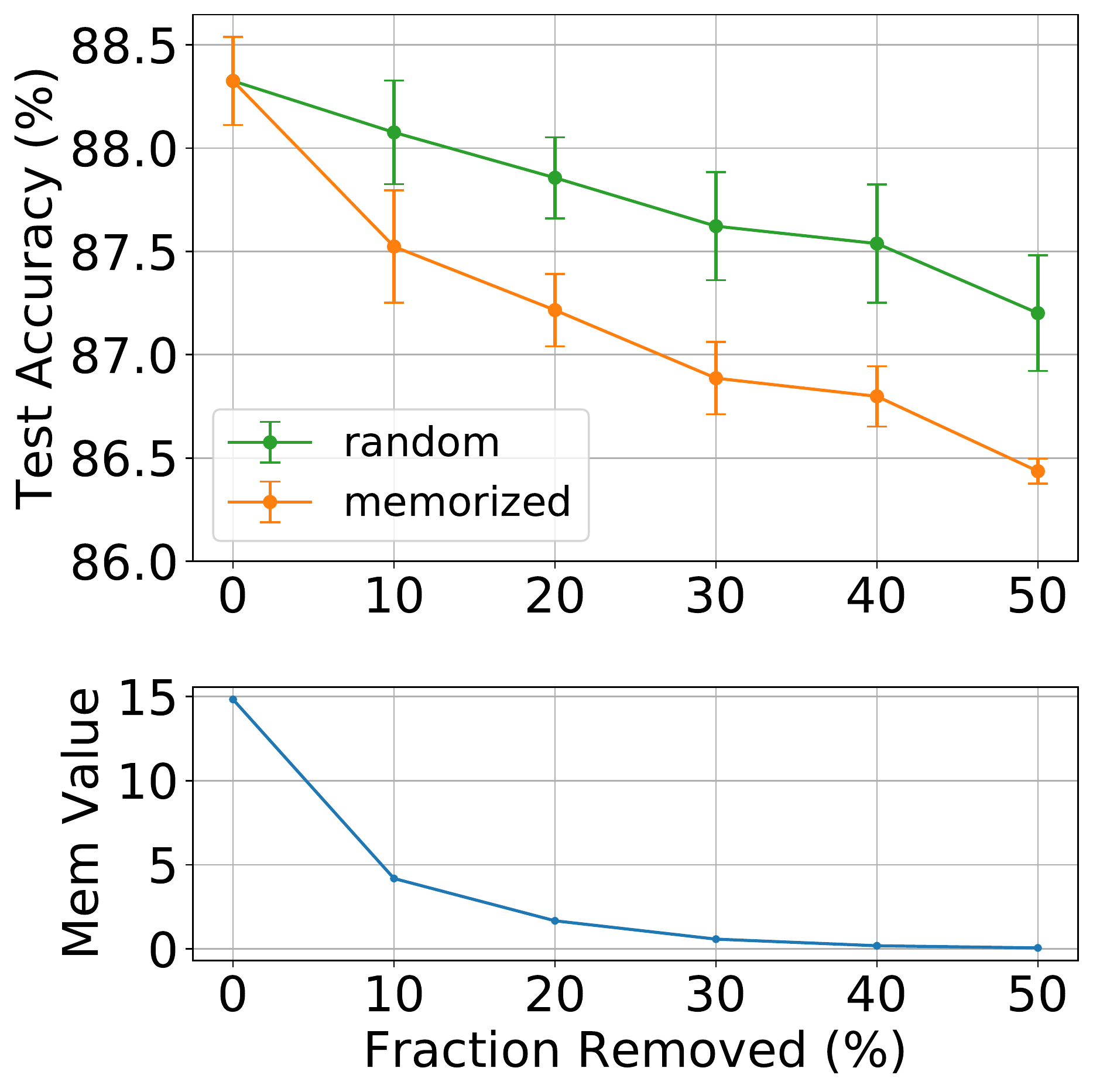}}
\subfigure[Yahoo! Answers]{
\label{Fig.Marginal.2}
\includegraphics[width=0.235\textwidth]{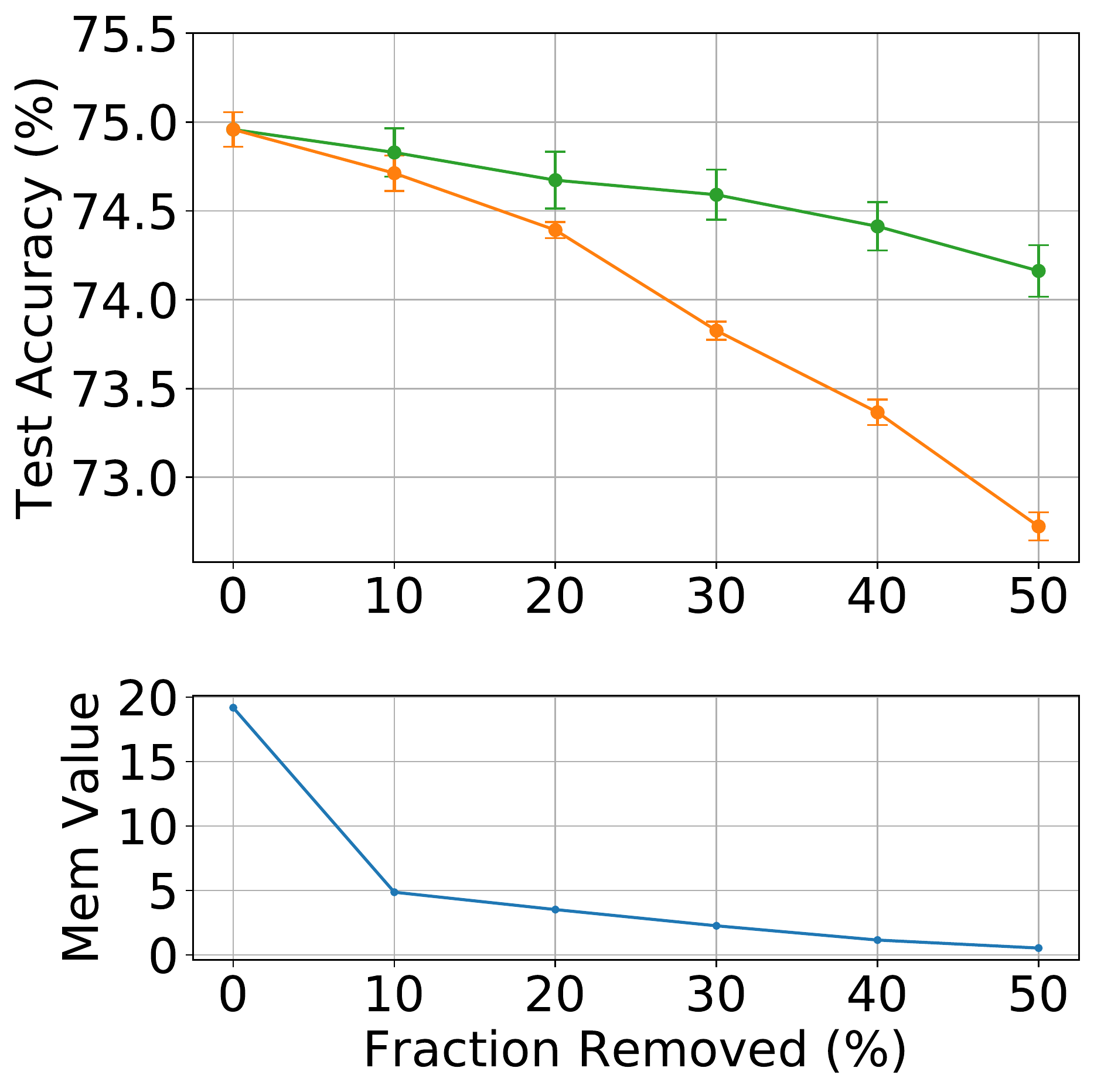}}
\subfigure[SNLI]{
\label{Fig.Marginal.3}
\includegraphics[width=0.235\textwidth]{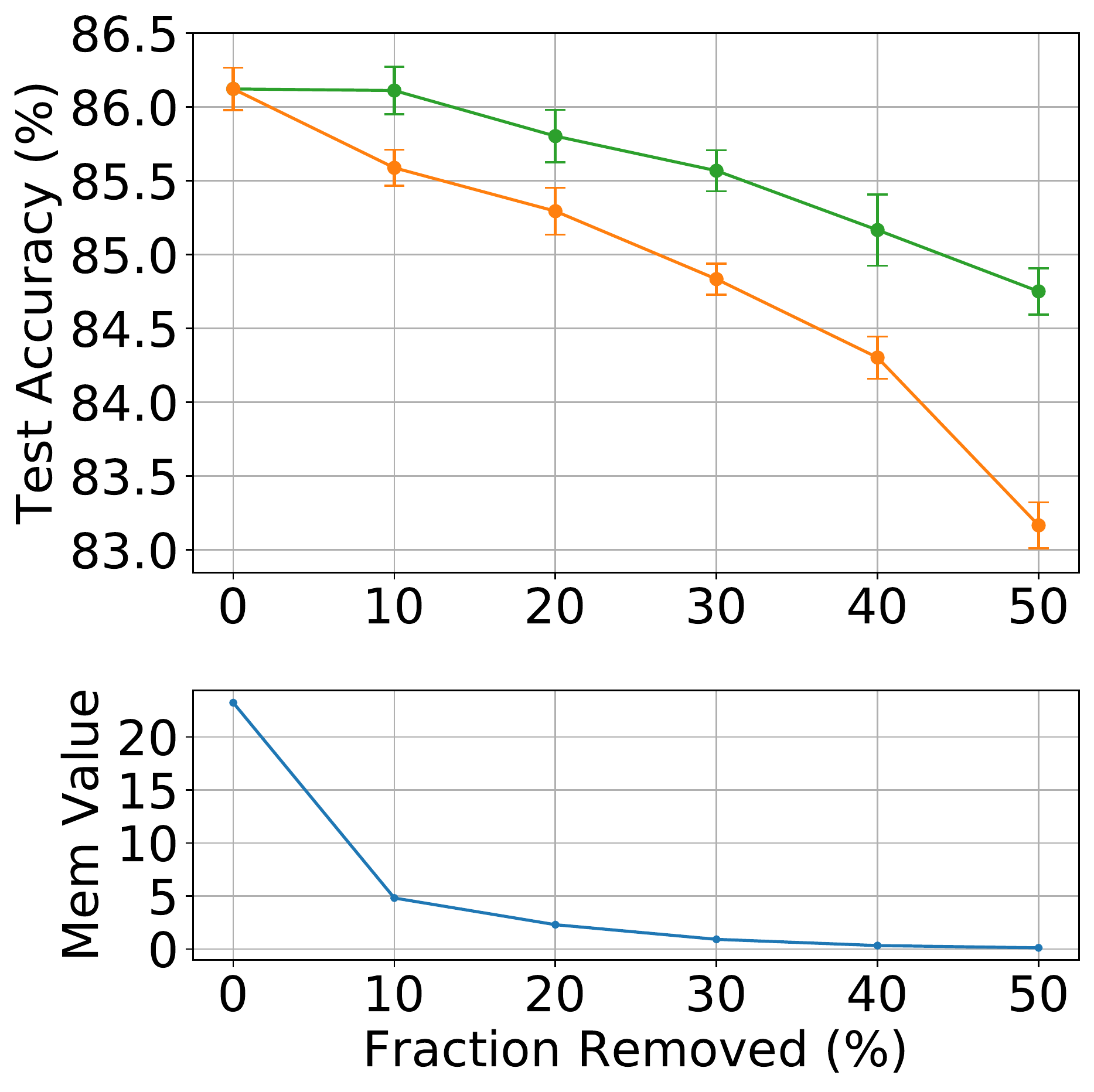}}
\subfigure[CIFAR-10]{
\label{Fig.Marginal.4}
\includegraphics[width=0.235\textwidth]{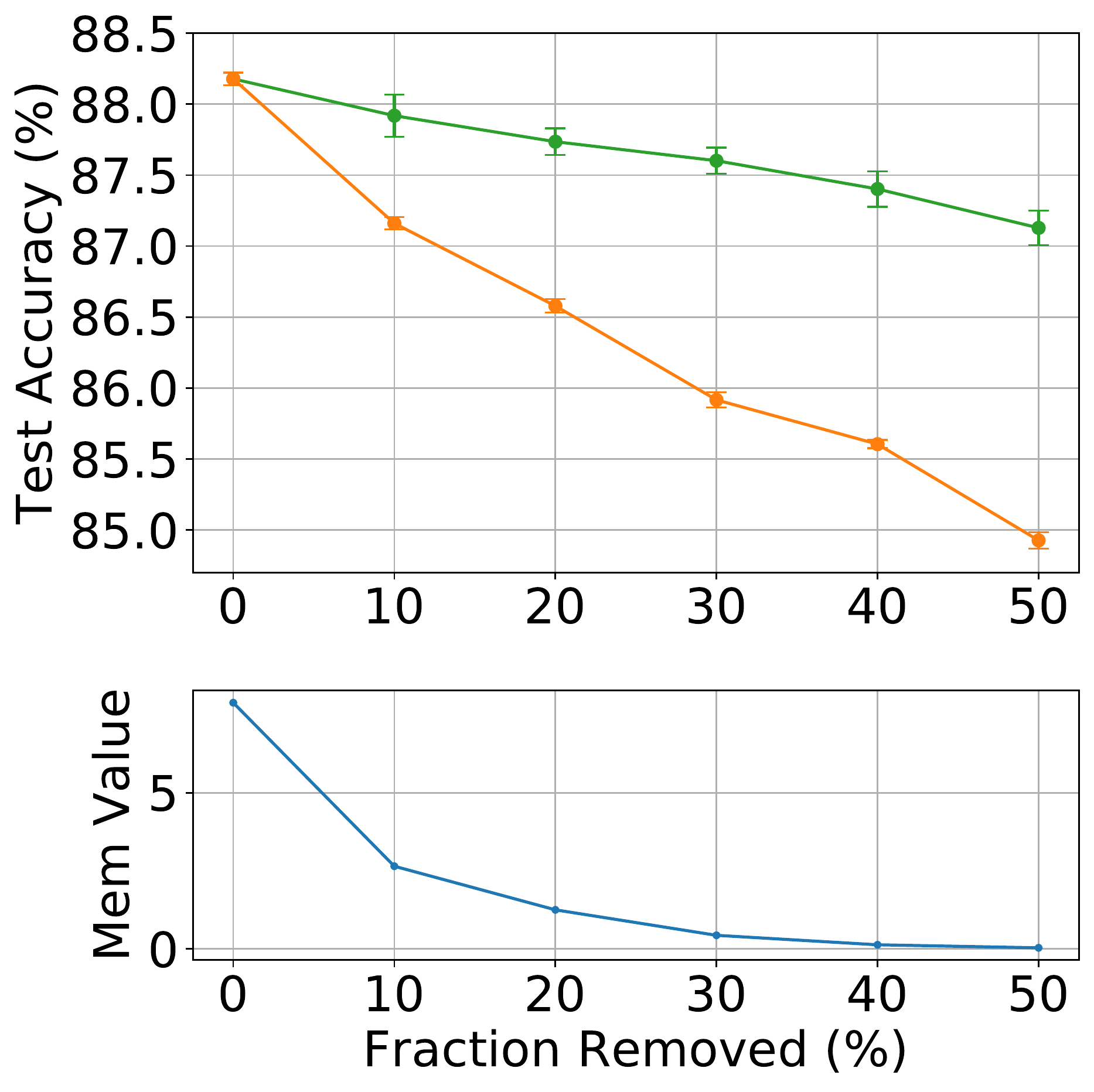}}

\caption{For each dataset, the top figure shows the test accuracy after we remove the top-$X\%$ 
memorized training instances or the same number of randomly selected training instances.
The test accuracy is averaged over 5 runs of retraining with different random seeds, and standard deviation is shown with the bars.
The bottom figure shows the lowest memorization score of the top-$X\%$ of the memorized training instances.
}
\label{Fig.Marginal}
\end{figure*}
\begin{figure*}[t]
\centering
\subfigure[SST-2]{
\label{Fig.Eval_attr.1}
\includegraphics[width=0.235\textwidth]{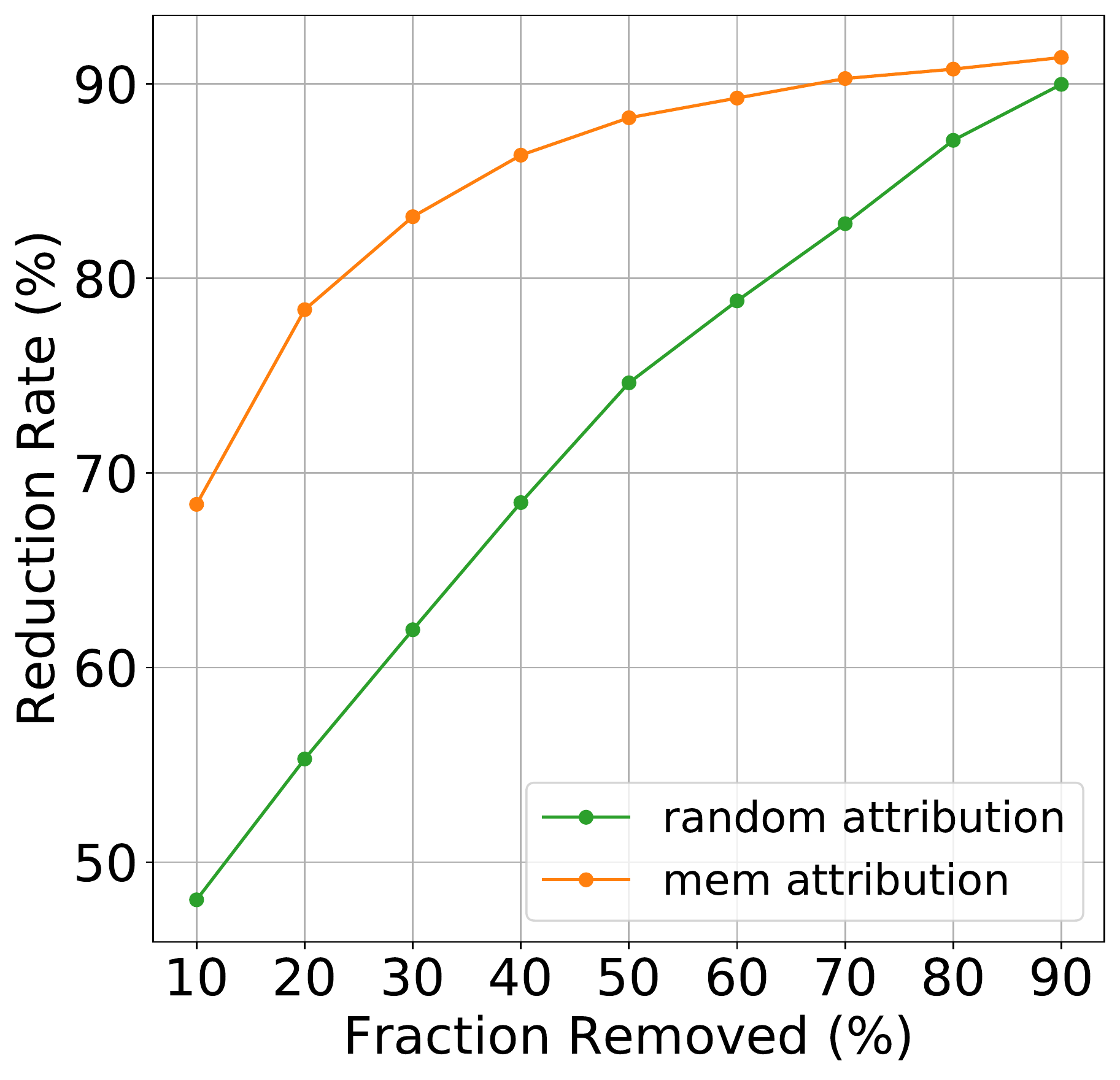}}
\subfigure[Yahoo! Answers]{
\label{Fig.Eval_attr.2}
\includegraphics[width=0.235\textwidth]{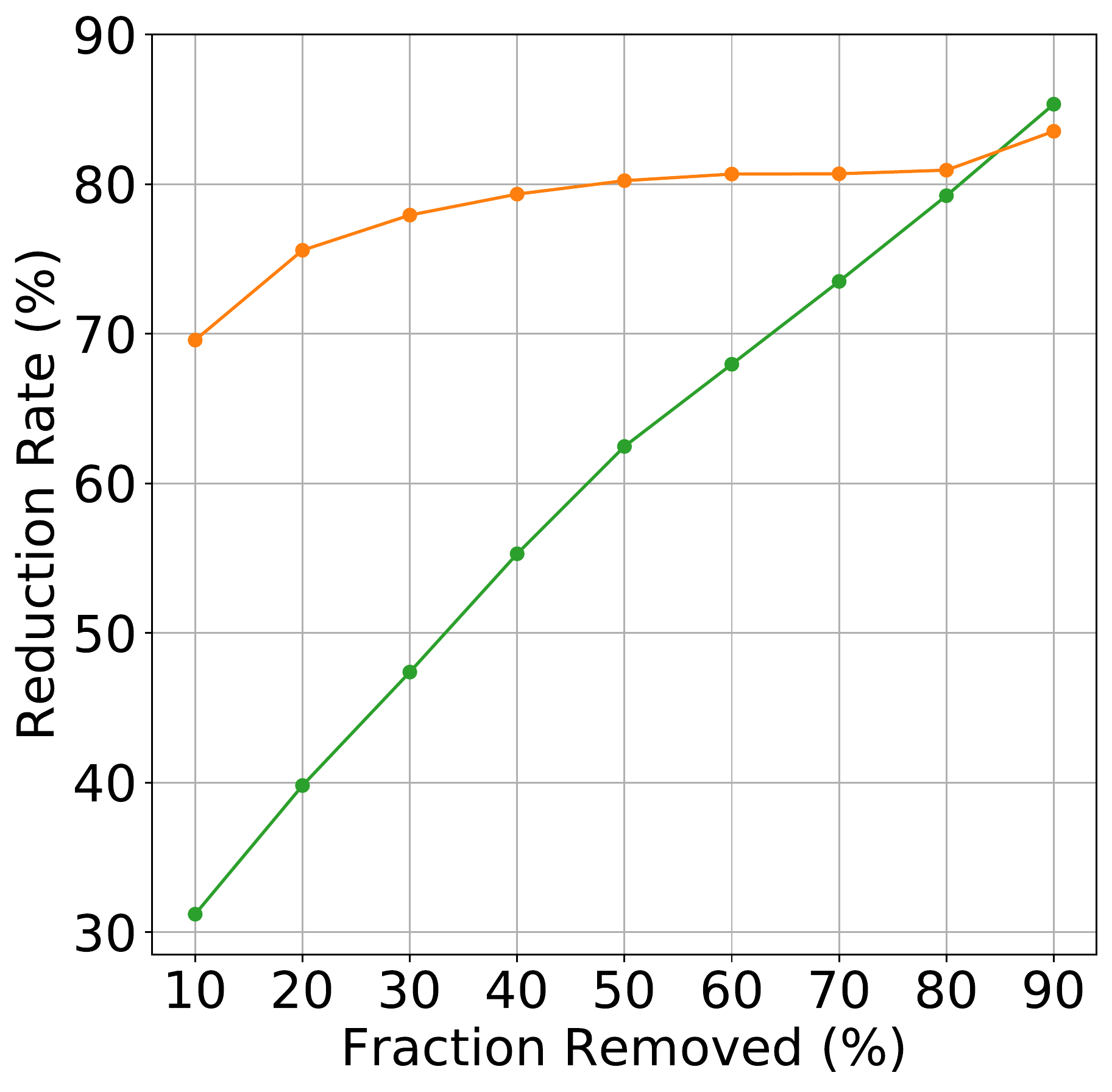}}
\subfigure[CIFAR-10]{
\label{Fig.Eval_attr.3}
\includegraphics[width=0.235\textwidth]{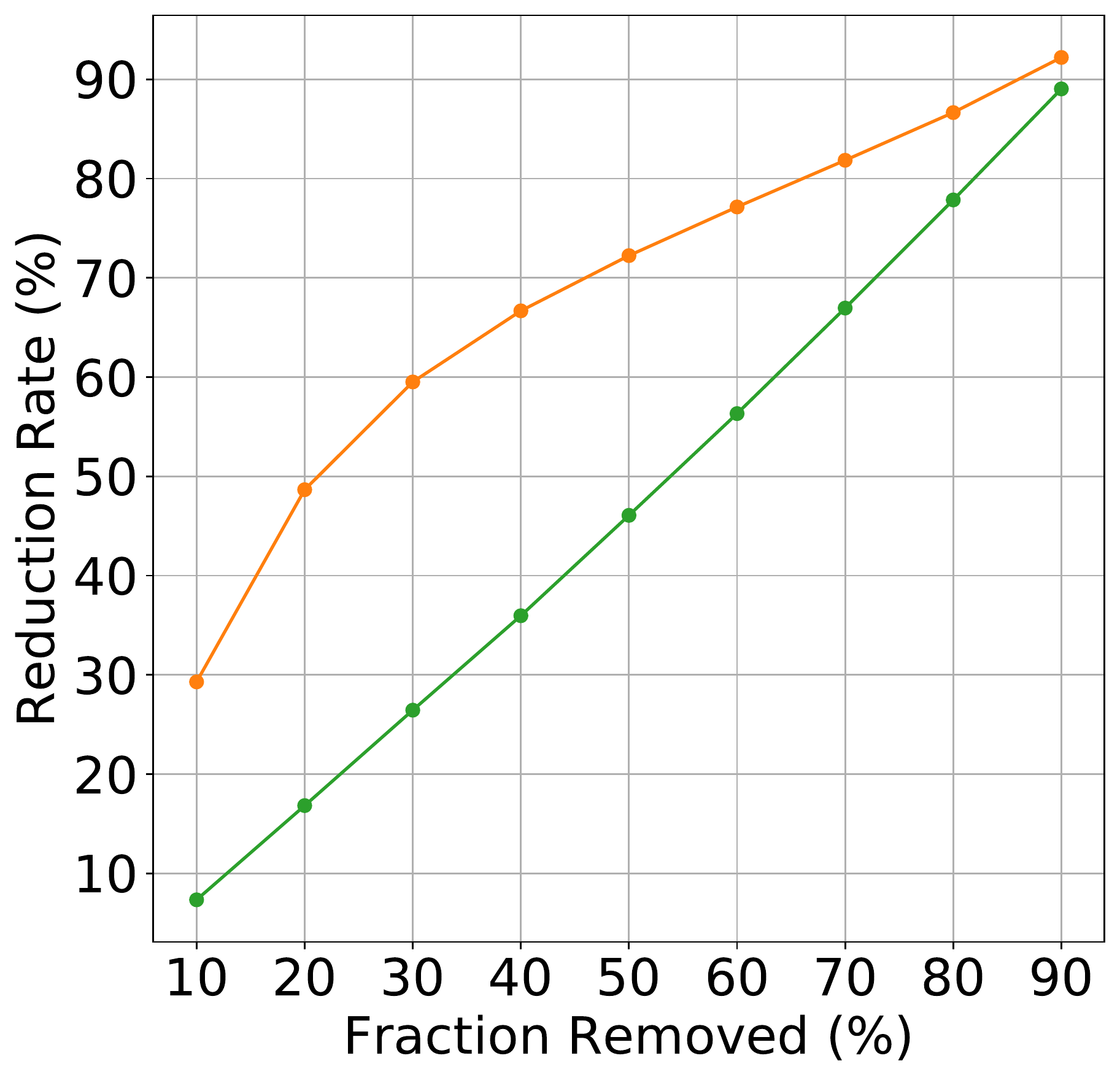}}

\caption{For each dataset, the top figure shows the reduction rate of removing the top-$k\%$ memorized tokens and of removing the same number of randomly selected training tokens. 
}
\label{Fig.Eval_attr}
\end{figure*}

In the second set of experiments, we check whether memorizing those training instances with the highest memorization scores leads to better performance on the unseen test data.
To do so, we compare the performance of the model on test data when top-ranked memorized training instances are removed during training versus the performance when the same number of \emph{randomly} selected training instances are removed.
If memorization is beneficial for the test data, then we would expect to see larger performance drop when top-ranked memorized training instances are removed than when random training instances are removed.
Therefore, the amount of performance drop represents the marginal effect of the memorized instances on the test accuracy.
We show the test accuracy in Figure~\ref{Fig.Marginal} when $X\%$ of the training instances are removed, where we set $X$ to a few different values.
We re-train the model 5 times and show the average test accuracy as well as the standard deviation.
We also show the lowest absolute memorization score of the top-$X\%$ of training instances in Figure~\ref{Fig.Marginal}.
For reference, here we also use CIFAR-10 to verify that our self-influence estimation using the influence function works similarly to the influence estimator used by \newcite{feldman2020neural}.

We can observe the following from Figure~\ref{Fig.Marginal}:
(1) On CIFAR-10~(Figure~\ref{Fig.Marginal.4}), we see that clearly the test accuracy drops more significantly when top-ranked memorized training instances instead of random training instances are removed.
Because \newcite{feldman2020neural} reported the same observation, this suggests that our memorization score based on the influence function proposed by \newcite{koh2017understanding} works similarly to the memorization estimator used by \newcite{feldman2020neural}.
This verifies the reliability of our memorization scoring function.
(2) On SST-2, Yahoo! Answers and SNLI, we can see that consistently when the same percentage of training instances are removed, removing top-ranked memorized instances has a clearly bigger impact on the test accuracy compared with removing random instances.
For example, on SST-2, the marginal utility of the top-30\% memorized training example is about 1.44 percentage points (vs. 0.70 percentage points for random subset of 30\% of training examples).

This verifies that on SST-2, Yahoo! Answers and SNLI, memorizing those training instances could help improve the performance on the test data.

\subsection{Evaluating Memorization Attribution}

In this section, we evaluate whether our memorization attribution method is faithful, i.e., whether it indeed picks up tokens that have higher self-influence.

Intuitively, if the memorization attribution method detects those memorized tokens in a training instance faithfully, then removing these tokens in that instance should result in a lower influence $\mathcal{I}$ of the perturbed instance on its original form (details to be given in Appendix~\ref{appendix:memorizaton_score}).
We therefore define a metric called Reduction Rate 
as follows:

\begin{equation}
    \begin{small}
    \frac{1}{|\mathcal{Z}|} \sum_{z \in \mathcal{Z}}{\frac{\mathcal{I}(z, z) - \mathcal{I}(z^{\setminus \text{attr}}, z)}{\mathcal{I}(z, z)}},
    \end{small}
\end{equation}
where $\mathcal{Z}$ is the set of top memorized training instances and $z^{\setminus \text{attr}}$ is the perturbed input where the top-$k\%$ memorized tokens are replace by the baseline token \texttt{[MASK]}. 
We can see that this Reduction Rate measures how much self-influence has been reduced after the top-memorized tokens are replaced with \texttt{[MASK]}.\footnote{We consider only top-$10\%$ memorized instances due to computation constraints.}

Figure~\ref{Fig.Eval_attr} demonstrates the significant effect of the removal of the top-memorized tokens from the top-memorized training instances. 
One could ask whether this effect is solely due to the input perturbation. 
To answer this question we include in the comparison the reduction rate of random attribution, i.e., we randomly remove some tokens from the training instances. 
We can see that removing tokens picked up by our memorization attribution method results in a much larger Reduction Rate until almost 90\% of the tokens are removed.
This result suggests that our memorization attribution method can indeed identify those tokens in a training instance that have high self-influence on that instance.

\subsection{Examples of Memorization Attribution}
\begin{table*}
\centering
\begin{small}
\begin{tabular}{ll}
\toprule
\textbf{Content} &\textbf{Label} \\
\midrule
\begin{CJK*}{UTF8}{gbsn}
{\setlength{\fboxsep}{0pt}\colorbox{white!0}{\parbox{0.8\textwidth}{
\colorbox{red!12.894462}{\strut starts} \colorbox{red!12.917763}{\strut out} \colorbox{red!5.3102994}{\strut with} \colorbox{red!63.996666}{\strut tremendous} \colorbox{red!48.31113}{\strut promise} \colorbox{red!13.504386}{\strut introducing} \colorbox{blue!5.0589848}{\strut an} \colorbox{red!38.367382}{\strut intriguing} \colorbox{red!4.3122873}{\strut and} \colorbox{red!17.149902}{\strut alluring} \colorbox{blue!4.594395}{\strut premise} \colorbox{red!12.398098}{\strut only} \colorbox{red!7.877758}{\strut to} \colorbox{red!1.695578}{\strut fall} \colorbox{red!0.24889877}{\strut prey} \colorbox{red!3.2718382}{\strut to} \colorbox{red!2.8787727}{\strut a} \colorbox{blue!1.1627978}{\strut boatload} \colorbox{blue!0.22193834}{\strut of} \colorbox{blue!4.673734}{\strut screenwriting} \colorbox{red!2.4392288}{\strut cliches} \colorbox{red!9.992298}{\strut that} \colorbox{red!6.3446865}{\strut sink} \colorbox{blue!0.5801939}{\strut it} \colorbox{red!8.74855}{\strut faster} \colorbox{red!7.717117}{\strut than} \colorbox{red!6.4836526}{\strut a} \colorbox{blue!1.5193763}{\strut leaky} \colorbox{red!1.1205802}{\strut freighter} 
}}}\end{CJK*} & Neg \\ 
\begin{CJK*}{UTF8}{gbsn}
{\setlength{\fboxsep}{0pt}\colorbox{white!0}{\parbox{0.8\textwidth}{
\colorbox{red!0.47934538}{\strut mr} \colorbox{red!4.506913}{\strut wollter} \colorbox{red!22.591978}{\strut and} \colorbox{red!1.3662844}{\strut ms} \colorbox{red!2.648377}{\strut seldhal} \colorbox{red!25.234634}{\strut give} \colorbox{red!59.484055}{\strut strong} \colorbox{red!18.91915}{\strut and} \colorbox{red!37.11892}{\strut convincing} \colorbox{blue!4.8403187}{\strut performances} \colorbox{blue!0.60171646}{\strut but} \colorbox{blue!8.0474825}{\strut neither} \colorbox{red!1.3953624}{\strut reaches} \colorbox{blue!1.7878284}{\strut into} \colorbox{blue!1.8901501}{\strut the} \colorbox{red!3.5680861}{\strut deepest} \colorbox{red!2.159412}{\strut recesses} \colorbox{red!2.4432347}{\strut of} \colorbox{red!1.3794377}{\strut the} \colorbox{red!0.74099696}{\strut character} \colorbox{red!5.498272}{\strut to} \colorbox{red!8.712304}{\strut unearth} \colorbox{red!2.9411633}{\strut the} \colorbox{red!1.7249041}{\strut quaking} \colorbox{red!3.120376}{\strut essence} \colorbox{red!4.384714}{\strut of} \colorbox{red!6.2231283}{\strut passion} \colorbox{blue!2.9908366}{\strut grief} \colorbox{red!5.6047273}{\strut and} \colorbox{blue!4.6749907}{\strut fear} 
}}}\end{CJK*} & Neg \\ 
\begin{CJK*}{UTF8}{gbsn}
{\setlength{\fboxsep}{0pt}\colorbox{white!0}{\parbox{0.8\textwidth}{
\colorbox{red!8.909469}{\strut this} \colorbox{red!33.77937}{\strut is} \colorbox{red!18.624908}{\strut a} \colorbox{red!55.578136}{\strut monumental} \colorbox{red!41.262417}{\strut achievement} \colorbox{red!5.3731484}{\strut in} \colorbox{blue!4.1069193}{\strut practically} \colorbox{red!4.6076636}{\strut every} \colorbox{red!1.3940359}{\strut facet} \colorbox{red!1.8146495}{\strut of} \colorbox{blue!45.811497}{\strut inept} \colorbox{red!9.2413}{\strut filmmaking} \colorbox{red!4.2215915}{\strut joyless} \colorbox{blue!16.374537}{\strut idiotic} \colorbox{blue!9.571635}{\strut annoying} \colorbox{blue!1.9741886}{\strut heavy} \colorbox{red!0.637793}{\strut handed} \colorbox{red!0.52921695}{\strut visually} \colorbox{red!5.2619357}{\strut atrocious} \colorbox{red!1.8014765}{\strut and} \colorbox{red!6.4284563}{\strut often} \colorbox{blue!6.3127823}{\strut downright} \colorbox{blue!12.012668}{\strut creepy} 
}}}\end{CJK*} & Neg \\ 
\midrule
\begin{CJK*}{UTF8}{gbsn}
{\setlength{\fboxsep}{0pt}\colorbox{white!0}{\parbox{0.8\textwidth}{
\colorbox{red!6.2263393}{\strut the} \colorbox{red!10.334582}{\strut director} \colorbox{red!15.317906}{\strut mark} \colorbox{red!9.5696945}{\strut pellington} \colorbox{red!8.6223}{\strut does} \colorbox{red!9.663196}{\strut a} \colorbox{blue!26.770805}{\strut terrific} \colorbox{red!12.171555}{\strut job} \colorbox{red!13.468639}{\strut conjuring} \colorbox{red!4.5437455}{\strut up} \colorbox{red!2.2544625}{\strut a} \colorbox{red!7.7666454}{\strut sinister} \colorbox{red!8.245412}{\strut menacing} \colorbox{red!1.4016712}{\strut atmosphere} \colorbox{blue!3.835258}{\strut though} \colorbox{red!56.419003}{\strut unfortunately} \colorbox{red!11.357104}{\strut all} \colorbox{red!8.11032}{\strut the} \colorbox{red!2.7061648}{\strut story} \colorbox{red!1.1618927}{\strut gives} \colorbox{red!1.1866478}{\strut us} \colorbox{red!6.651921}{\strut is} \colorbox{red!3.8103228}{\strut flashing} \colorbox{red!7.5579667}{\strut red} \colorbox{red!3.975449}{\strut lights} \colorbox{red!19.46616}{\strut a} \colorbox{red!12.293593}{\strut rattling} \colorbox{red!3.4819107}{\strut noise} \colorbox{red!14.094279}{\strut and} \colorbox{red!23.225025}{\strut a} \colorbox{red!22.66697}{\strut bump} \colorbox{red!4.0177054}{\strut on} \colorbox{red!3.4850764}{\strut the} \colorbox{red!6.7588253}{\strut head} 
}}}\end{CJK*} & Pos \\ 
\begin{CJK*}{UTF8}{gbsn}
{\setlength{\fboxsep}{0pt}\colorbox{white!0}{\parbox{0.8\textwidth}{
\colorbox{red!4.916347}{\strut this} \colorbox{blue!2.0631433}{\strut is} \colorbox{blue!0.3493431}{\strut a} \colorbox{blue!33.19824}{\strut fascinating} \colorbox{red!1.4569104}{\strut film} \colorbox{blue!5.9088373}{\strut because} \colorbox{red!3.065613}{\strut there} \colorbox{red!2.3782227}{\strut is} \colorbox{red!86.87165}{\strut no} \colorbox{red!5.1966734}{\strut clear} \colorbox{red!6.7440004}{\strut cut} \colorbox{blue!0.4221037}{\strut hero} \colorbox{red!0.30667293}{\strut and} \colorbox{red!22.927729}{\strut no} \colorbox{red!1.9039879}{\strut all} \colorbox{red!2.5470462}{\strut out} \colorbox{red!7.4542055}{\strut villain} 
}}}\end{CJK*} & Pos \\ 
\begin{CJK*}{UTF8}{gbsn}
{\setlength{\fboxsep}{0pt}\colorbox{white!0}{\parbox{0.8\textwidth}{
\colorbox{red!12.89524}{\strut the} \colorbox{red!12.791196}{\strut film} \colorbox{red!4.9860806}{\strut is} \colorbox{red!11.576382}{\strut reasonably} \colorbox{blue!13.122864}{\strut entertaining} \colorbox{red!19.2303}{\strut though} \colorbox{red!4.796516}{\strut it} \colorbox{red!30.859413}{\strut begins} \colorbox{red!0.68088317}{\strut to} \colorbox{red!24.619133}{\strut drag} \colorbox{red!11.582205}{\strut two} \colorbox{red!8.557351}{\strut thirds} \colorbox{red!13.219282}{\strut through} \colorbox{red!0.97397685}{\strut when} \colorbox{red!6.355854}{\strut the} \colorbox{red!38.55148}{\strut melodramatic} \colorbox{red!6.1039376}{\strut aspects} \colorbox{red!19.049177}{\strut start} \colorbox{red!3.149822}{\strut to} \colorbox{red!23.863108}{\strut overtake} \colorbox{red!10.206303}{\strut the} \colorbox{red!7.71556}{\strut comedy} 
}}}\end{CJK*} & Pos \\ 

\bottomrule
\end{tabular}
\end{small}
\caption{The top-3 memorized training instances for each class from SST-2.
Highlighted words are those with high attribution values (red for positive memorization attribution and blue for negative memorization attribution) as computed by our memorization attribution method.
}
\label{table.vis.sst.attr}
\end{table*}

\begin{figure*}
\centering
\includegraphics[width=0.9\textwidth]{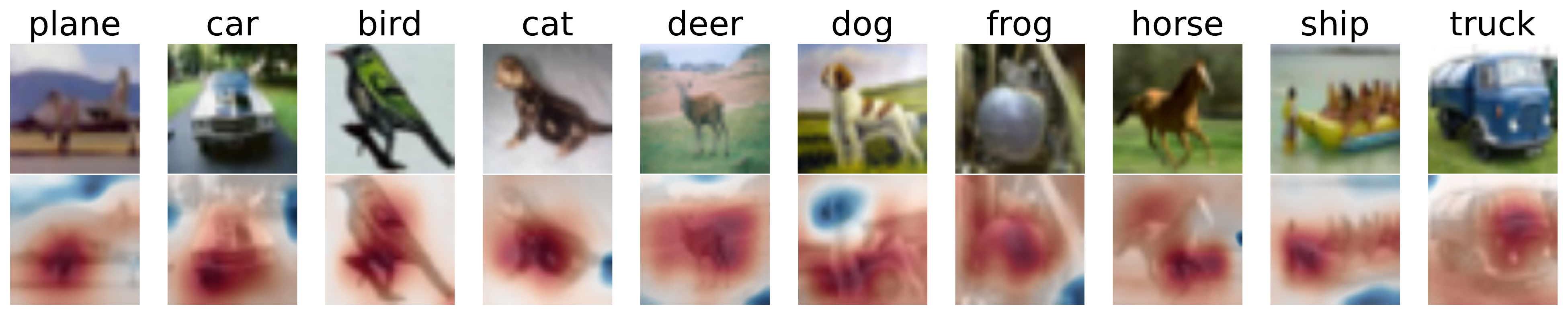}
\caption{The top-1 memorized training instance for each class from CIFAR-10.
Highlighted patches are those having high attribution values (red for positive memorization attribution and blue for negative memorization attribution) as computed by our memorization attribution method.
}
\label{Fig.CIFAR}
\end{figure*}
To better understand why certain training instances are memorized, we apply our memorization attribution method to SST-2, Yahoo! Answers and CIFAR-10.
We do not discuss our memorization attribution method applied to the NLI task because we find that it is not easy to interpret the results.
In some other studies~(e.g., \newcite{han-etal-2020-explaining}), people have also reported different behaviours of NLI from tasks relying on shallow features such as sentiment classification and topic-based text classification.

We find that on SST-2, Yahoo! Answers and CIFAR-10, in most cases our memorization attributions are easy to be interpreted by humans. 
In particular, without any cherry-picking, we select those instances with the highest memorization scores to present.
We find that interestingly, for both SST-2 and CIFAR-10, the trained deep learning model tends to memorize those parts of an instance that are negatively correlated with the class label of that instance, as shown in Table~\ref{table.vis.sst.attr} and Figure~\ref{Fig.CIFAR}.\footnote{For Yahoo! Answers, because each instance is long, due to the space limit, we show the memorization attributions in the Appendix~\ref{appendix:attribution}.}
On SST-2, for example, the model needs to memorize positive phrases such as ``tremendous promise'' and ``intriguing and alluring'' that show up in an overall negative instance.
On CIFAR-10, we observe that for images that are easily mis-classified, the model memorizes those pixels that are associated with the wrong class label, or in other words, pixels that are negatively correlated with the correct class label. For example, the ``cat'' image shown in Figure~\ref{Fig.CIFAR} looks like a frog. The model memorizes those pixels (shown in red) around the tummy of the cat because those pixels make the image look like a frog image. Similarly, in the ``dog'' image, which looks like a horse, the memorized pixels (shown in red) are around the body of the dog, and these pixels make the image look like a horse image. 
On the other hand, the dog's head in this image, which is a typical dog's head, has negative memorization attribution scores, which means it does not need to be memorized.

Given the interesting results above, we believe that model developers can gain insights about what a model finds hard to learn from other training instances (and thus has to memorize), and model developers can subsequently take actions like up-weighting memorized instances or collecting similar data to improve the performance on certain subpopulations if desired.

\section{Related Work}

\paragraph{The long-tail theory:}
The long-tail theory proposed by \newcite{feldman2020does} 
is relatively new and has not been systematically validated in NLP.
Our work 
is the first to empirically check the validity of this theory on NLP tasks.
\newcite{raunak-etal-2021-curious} used the long-tail theory to explain hallucinations under source perturbations in Neural Machine Translation.
They assume the theory holds in NMT rather than validating the theory itself as we do. 
\newcite{kong2021understanding} investigated the memorization phenomenon for Variational Auto-Encoder also via self-influence.

\paragraph{Memorization vs. generalization:}
It is well-known that deep learning models possess strong capabilities to memorize training instances \cite{zhang2017understanding, arpit2017closerlook}.
In the context of NLP, \newcite{li-wisniewski-2021-neural} showed that BERT is more likely to memorize shallow patterns from the training data rather than uncover abstract properties.
Some recent work has tried to combine interpolation methods with deep learning methods to generalize via memorization~\cite{khandelwal2019generalization, khandelwal2020nearest}.
However, 
previous work using interpolation methods did not explain why memorization is necessary in the first place. 
Our work follows the long-tail theory that views %
memorization as beneficial to generalization when the data follows a certain type of long-tail distribution. 
There has also been some work studying ``forgetting,'' which is related to memorization~\cite{toneva2018empirical, yaghoobzadeh-etal-2021-increasing}.
However, in this paper we do not study this ``forgetting'' phenomenon.

\paragraph{Influence functions:}

Influence functions have been studied for large-scale deep neural networks by \newcite{koh2017understanding} and gained much attention in recent years.
In the context of 
NLP, \newcite{han-etal-2020-explaining} explored the usage of influence functions to explain model predictions and unveil data artifacts.
\newcite{meng2020pair} proposed a combination of gradient-based methods and influence functions to examine training history and test stimuli simultaneously.
Our work, however, adopts influence function as a tool to measure memorization.

\section{Conclusions}
In this paper, we empirically examine a recently proposed long-tail theory in the context of NLP.
We use sentiment classification, natural language inference and text classification to check the validity of the long-tail theory in NLP.
We also propose a memorization attribution method to reveal which parts of an instance are being memorized. 
There are two major takeaway messages: (1) Our experiments empirically validated the long-tail theory on the three NLP datasets, showing that memorization is important for generalization,
offers an alternative view and helps NLP researchers to see the value of memorization.
(2) Our attribution method can be a tool to help model developers better understand the memorization behaviours of a model and possibly further improve the model.

\section{Ethical Considerations}
Our work empirically validated the long-tail theory in the context of NLP, offering an alternative view to the relationship between memorization and generalization.
This will help NLP researchers see the value of memorization. 
However, previous work showed that neural networks can be vulnerable to privacy attacks such as membership inference attacks because these models are able to memorize training instances, and sometimes sensitive private information may be contained in the training instances~\cite{shokri2017membership, zhang2017understanding, feldman2020neural}. 
Thus, there is a trade-off between the accuracy of a model and the privacy of the data.
In other words, although memorization can help reduce generalization error (as we showed in this paper), it also increases the vulnerability of the system to privacy attacks, which raises ethical concerns.

The computation of influence functions used in our work is massive because of the computation of inverting the hessian matrices. 
To reduce the computation costs, i.e., power consumption, we may adopt other influence estimators like TracIn~\cite{pruthi2020estimating}, which is hessian-free and thus faster.

\section*{Acknowledgment}

This research is supported by the Singapore Ministry of Education (MOE) Academic Research Fund (AcRF) Tier 1 grant.

\bibliography{anthology,custom}
\bibliographystyle{acl_natbib}

\clearpage
\appendix

\section{Derivation of the Memorization Scores}
\label{appendix:memorizaton_score}

For clarity, here we repeat the derivation of Influence Functions by \newcite{koh2017understanding} and provide 
self-influence functions as its special case.
Note that self-influence functions are used as our memorization scores.

Given training points $z_1, ..., z_n$, where $z_i = (x_i, y_i)$, $x_i$ is the observation and $y_i$ is the label, we train a predictor via minimizing the empirical risk $R(\theta) \overset{\text{def}}{=} \frac{1}{n} \sum^n_{i=1}{L(z_i, \theta)}$ to pick parameters $\theta \in \Theta$. 
I.e., the optimal parameters are obtained by $\hat{\theta} = \arg\min_{\theta \in \Theta} R(\theta)$.
We assume that $R$ is twice-differentiable and strongly convex.

i.e.,
\begin{equation}
   H_{\hat{\theta}} \overset{\text{def}}{=} \nabla^2{R(\hat{\theta}}) = \frac{1}{n} \sum^n_{i=1}{\nabla^2_\theta{L(z_i, \hat{\theta})}}
\label{eq: Hessian}
\end{equation}
exists and is positive definite. 
This guarantees the existence of $H^{-1}_{\hat{\theta}}$, which we will use in the following derivation. 

The high-level idea of Influence Functions is to approximate leave-one-out retraining, which corresponds to a removing operation, via computing the parameter change if $z$ were up-weighted or down-weighted by some small amount $\epsilon$.

If we up-weight the training point $z$, the perturbed parameters $\hat{\theta}_{\epsilon, z}$ can be written as
\begin{equation}
    \hat{\theta}_{\epsilon, z} = \arg\min_{\theta \in \Theta} \left( R(\theta) + \epsilon L(z, \theta) \right).
\label{eq: argmin}
\end{equation}

Consider the parameter change $\Delta_{\epsilon} = \hat{\theta}_{\epsilon, z} - \hat{\theta}$, and note that, as $\hat{\theta}$ does not depend on $\epsilon$, the quantity we want to compute can be written in terms of it:
\begin{equation}
    \frac{d\hat{\theta}_{\epsilon, z}}{d\epsilon} = \frac{d \Delta_{\epsilon}}{d\epsilon}.
\label{eq: seek}
\end{equation}

Since $\hat{\theta}_{\epsilon, z}$ is a minimizer of Eqn~\ref{eq: argmin}, let us examine its first-order optimality condition:
\begin{equation}
    0 = \nabla{R(\hat{\theta}_{\epsilon, z})} + \epsilon \nabla{L(z, \hat{\theta}_{\epsilon, z})}.
\end{equation}
Let us define $f(\theta)$ to be $\nabla{R(\theta)} + \epsilon \nabla{L(z, \theta)}$.

Next, since $\hat{\theta}_{\epsilon, z} \rightarrow \hat{\theta}$ as $\epsilon \rightarrow 0$, we perform a Taylor expansion on $f(\hat{\theta}_{\epsilon, z})$.
Given Taylor's Formula $f(\theta+\Delta \theta) = f(\theta) + f'(\theta) \Delta \theta + o(\Delta \theta)$,
we have:
\begin{equation}
\begin{aligned}
    0 
    &= f(\hat{\theta}_{\epsilon, z}) \\
    &= f(\hat{\theta} + \Delta_{\epsilon}) \\
    &\approx f(\hat{\theta}) +  f'(\hat{\theta})\Delta_{\epsilon} \\
    &= [\nabla{R(\hat{\theta})} + \epsilon \nabla{L(z, \hat{\theta})}] \\
    & \quad +[\nabla^2{R(\hat{\theta})} + \epsilon \nabla^2{L(z, \hat{\theta})}]\Delta_{\epsilon},
\end{aligned}
\end{equation}
where we have dropped the term $o(\|{\Delta_{\epsilon}}\|)$.

Solving for $\Delta_{\epsilon}$, we get
    $\Delta_{\epsilon} \approx -[\nabla^2{R(\hat{\theta})} + \epsilon \nabla^2{L(z, \hat{\theta})}]^{-1} [\nabla{R(\hat{\theta})} + \epsilon \nabla{L(z, \hat{\theta})}]$.

Since $\hat{\theta}$ minimizes $R$, we have $\nabla R(\hat{\theta}) = 0$. 
Then we have:
\begin{equation}
\begin{aligned}
    \Delta_{\epsilon} &\approx -[\nabla^2{R(\hat{\theta})} + \epsilon \nabla^2{L(z, \hat{\theta})}]^{-1} \epsilon \nabla{L(z, \hat{\theta})}.
\end{aligned}
\end{equation}

Referring to \cite{henderson1981deriving}, we have: 
\begin{equation}
\begin{aligned}
    (A+B)^{-1} 
    &= (I + A^{-1}B)^{-1}A^{-1} \\ 
    &= A^{-1} - A^{-1}B(I + A^{-1}B)^{-1}A^{-1} \\ 
    &= A^{-1} - A^{-1}B(A+B)^{-1},
\end{aligned}
\end{equation}
which only requires $A$ and $A+B$ to be nonsingular matrix. 
As mentioned above, the matrices that we are considering are positive definite. The determinant of a positive definite matrix is always positive, so a positive definite matrix is always nonsingular.

Substituting $A=\nabla^2{R(\hat{\theta})}$ and $B=\epsilon \nabla^2{L(z, \hat{\theta})}$ and dropping $o(\epsilon)$ terms, we have 
\begin{equation}
    \Delta_{\epsilon} \approx -\nabla^2{R(\hat{\theta})}^{-1} \nabla{L(z, \hat{\theta})} \epsilon.
\end{equation}

Combining with Eqn~\ref{eq: Hessian} and Eqn~\ref{eq: seek}, we conclude that:
\begin{equation}
\begin{aligned}
    \frac{d \hat{\theta}_{\epsilon, z}}{d \epsilon} \bigg|_{\epsilon=0} 
    &= -H^{-1}_{\hat{\theta}} \nabla L(z, \hat{\theta}). \\
\end{aligned}
\end{equation}

We instead down-weight the training point $z$ to keep consistency with our memorization attribution method introduced later, 
the perturbed parameters $\hat{\theta}_{\epsilon, -z}$ can be written as
\begin{equation}
    \hat{\theta}_{\epsilon, -z} = \arg\min_{\theta \in \Theta} \left( R(\theta) - \epsilon L(z, \theta)\right).
\end{equation}

It is easy to see that
\begin{equation}
\begin{aligned}
    \frac{d \hat{\theta}_{\epsilon, -z}}{d \epsilon} \bigg|_{\epsilon=0} 
    &= H^{-1}_{\hat{\theta}} \nabla L(z, \hat{\theta}). \\
\end{aligned}
\end{equation}

Next, we apply the chain rule to measure how down-weighting $z$ changes functions of $\hat{\theta}$.
\begin{equation}
\begin{aligned}
    \mathcal{I}(z, z_\text{test})
    &\overset{\text{def}}{=} \frac{d F(y_\text{test}, x_\text{test}; \hat{\theta}_{\epsilon, -z})}{d \epsilon} \bigg|_{\epsilon=0} \\
    & = \nabla_{\theta}F(y_\text{test}, x_\text{test}; \hat{\theta})^{\top}\frac{d \hat{\theta}_{\epsilon, -z}}{d \epsilon} \bigg|_{\epsilon=0} \\
    &= \nabla_{\theta}F(y_\text{test}, x_\text{test}; \hat{\theta})^{\top}H^{-1}_{\hat{\theta}}\nabla_{\theta}{L(z, \hat{\theta})},
\end{aligned}
\label{IF.original}
\end{equation}
where $F$ is usually the loss function.

While influence function is generally used to measure the influence of a training instance on a test instance, if we use it to measure the influence of a training instance on itself, i.e., to measure self-influence, then this self-influence corresponds to the general notion of memorization defined by \newcite{feldman2020does, feldman2020neural}. Based on this notion, we set $F$ as the negative estimated conditional probability $-P(y|x; \theta)$ and define the memorization score for a training instance $z$ as follows:
\begin{equation}
\begin{aligned}
    \mathcal{M}_{\text{remove}}(z) 
    & \overset{\text{def}}{=} -\frac{d P(y|x; \hat{\theta}_{\epsilon, -z})}{d \epsilon} \bigg|_{\epsilon=0} \\
    &= -\nabla_{\theta}P(y|x; \hat{\theta})^{\top}\frac{d \hat{\theta}_{\epsilon, -z}}{d \epsilon} \bigg|_{\epsilon=0} \\
    &= -\nabla_{\theta}P(y|x; \hat{\theta})^{\top}H^{-1}_{\hat{\theta}}\nabla_{\theta}{L(z, \hat{\theta})}.
\end{aligned}
\end{equation}

\section{Derivation of Memorization Attribution}
\label{appendix:memorizaton_attribution}

In order to better understand why an instance is memorized, we propose a fine-grained notion of memorization at ``feature'' level instead of instance level, i.e., to attribute the memorization score of an instance to its individual features.

To conduct attribution, a natural requirement is to introduce a baseline. Thus we first consider a variant of the Influence Functions that approximates the resulting effect of \emph{replacing} a training point $z$ with a baseline training point $z'$, which is similar to the perturbation-based influence by \newcite{koh2017understanding}.

The perturbed parameter $\hat{\theta}_{\epsilon, z_{\delta}, -z}$ can be written as:
\begin{equation}
   \hat{\theta}_{\epsilon, z', -z} = \arg\min_{\theta \in \Theta} \left( R(\theta) + \epsilon L(z', \theta) - \epsilon L(z, \theta) \right).
\end{equation}

Similar to the derivation shown the previous section, we can derive the following definition of a memorization score based on such perturbation:
\begin{equation}
\begin{aligned}
    \mathcal{M}_{\text{replace}}(z)
    &\overset{\text{def}}{=} -\frac{d P(y|x; \hat{\theta}_{\epsilon, z', -z})}{d \epsilon} \bigg|_{\epsilon=0} \\
    & = -\nabla_{\theta}P(y|x; \hat{\theta})^{\top}\frac{d \hat{\theta}_{\epsilon, z', -z}}{d \epsilon} \bigg|_{\epsilon=0} \\
    &= -s^{\top} \left(\nabla_{\theta}{L(z, \hat{\theta})} - \nabla_{\theta}{L(z', \hat{\theta})}\right),
\end{aligned}
\label{appendix.replace.1}
\end{equation}
where $s=H^{-1}_{\hat{\theta}}\nabla_{\theta}{P(y|x; \hat{\theta})}$.

We now show that $\mathcal{M}_{\text{replace}}(z)$ can be decomposed into a linear combination of scores, each corresponding to a single token in the input sequence.
For NLP applications, the input $x$ usually corresponds to an embedding matrix $\mathbf{X} \in \mathbb{R}^{N \times d}$ (where $N$ is the number of tokens and $d$ is the embedding dimensions).
Let us denote $\nabla_{\theta}{L\left((\cdot, y), \hat{\theta}\right)}$ as $g(\cdot)$ and consider the path integral along a straight line between $\mathbf{X}$ and $\mathbf{X}'$, yielding

\begin{equation}
\begin{aligned}
    g(\mathbf{X}) - g(\mathbf{X}') 
    &= H'(\mathbf{X} - \mathbf{X}'),
\end{aligned}
\label{appendix.gradient}
\end{equation}
where $H'=\left[\int^1_{\alpha = 0}{\frac{d g\left(\mathbf{X}' + \alpha (\mathbf{X} - \mathbf{X}') \right)}{d x} d\alpha} \right]$ and could be efficiently approximated by Riemann Sum as suggested by \newcite{sundararajan2017axiomatic}.

The reason of using path integral rather than the gradient at the input $\mathbf{X}$ is that a function's gradient may saturate
around the input and integrating along a path can alleviate this issue. 
As for the reasons of choosing a straight line between the input and the baseline,
first of all, it is obviously the simplest path. 
Besides, using a straight line allows the Integrated Gradients to meet the Symmetry-Preserving property. 
For more details, please check the original paper on IG~\cite{sundararajan2017axiomatic}.

Substituting Eqn~(\ref{appendix.gradient}) into Eqn~(\ref{appendix.replace.1}), we get

\begin{equation}
\begin{aligned}
    \mathcal{M}_{\text{replace}}(z) 
    &= -s^{\top} \left(g(\mathbf{X}) - g(\mathbf{X}')\right) \\
    &= -s^{\top} H' (\mathbf{X} - \mathbf{X}') \\    
    &= -r^{\top}(\mathbf{X} - \mathbf{X}') \\
    &= -\sum^{N}_{t=1}{ \sum^{d}_{l=1}{r_{t,l} (\mathbf{X}_{t, l} - \mathbf{X}'_{t, l})}},
\end{aligned}
\label{appendix.replace.2}
\end{equation}
where $r=H's$, which could be efficiently computed by the the hessian-vector product \cite{pearlmutter1994fast}.

\section{The Effect of Different Checkpoints}
\label{appendix:checkpoints}

Our self-influence-based memorization score is dependent on the model used to compute the influence function.
A model trained with different random seeds will have different self-influence values, so there is inherently some stochasticity in the measurement of influence or self influence.

To address this issue, on SST-2, we train the model using different random seeds to obtain three checkpoints and compute the corresponding memorization scores.
We found that the instance rankings produced by these different checkpoints are highly correlated, based on Spearman's Rank Correlation Coefficient, as shown in Table~\ref{table.correlation}. Thus, we only consider one checkpoint when computing the memorization scores.

\begin{table}[h]
    \centering
    \begin{tabular}{ll|l|l|l|}
    \multicolumn{2}{c}{} & \multicolumn{1}{c}{a} & \multicolumn{1}{c}{b} & \multicolumn{1}{c}{c} \\ \cline{3-5}
    \multirow{3}{*}{{\rotatebox[origin=c]{90}{seed}
        }} 
        & a &1.00 & 0.99 & 0.98  \\ \cline{3-5}
        & b &0.99 & 1.00 & 0.99 \\ \cline{3-5}
        & c &0.98 & 0.99 & 1.00 \\ \cline{3-5}
    \end{tabular}
    \caption{Spearman's rank correlation coefficients between different rankings of the training instances produced by different checkpoints of the trained model on SST-2.}
    \label{table.correlation}
\end{table}

\section{The Distribution of Positive Phrase Fraction}
\label{appendix:atypical}

For the task of sentiment classification, i.e., the experiments on SST-2, we hypothesize 
that a typical positive sentence should have a relatively high percentage of positive phrases and a typical negative sentence should have a relatively low percentage of positive phrases. 
Note that here we consider phrase-level sentiment instead of word-level sentiment because we want to take into account the negation phenomena such as the phrase ``not bad" expressing a positive sentiment although the word ``bad'' contains a negative sentiment.
To support our hypothesis, we conduct the following quantitative experiment. 

Given the phrase-level sentiment annotations provided by the SST-2 dataset~\cite{socher-etal-2013-recursive}, for every instance $z$, we count how many positive phrases and negative phrases it contains, respectively. 
Then, we turn the absolute counts into a relative fraction:
\begin{equation}
    \text{frac}(z)_{c} = \frac{\text{count}(z)_{c} + k}{\sum_{c' \in \{\text{neg}, \text{pos}\}}{ \left( \text{count}(z)_{c'} + k \right)}},
\end{equation}
where $\text{count}(z)_{c}$ is the number of phrases with sentiment $c$ in instance $z$,
and add-$k$-smoothing is used to avoid division by zero. Here $k$ is set as 0.01.

\begin{figure}
\centering
\subfigure[Negative Class]{
\label{Fig.Atypical.1}
\includegraphics[width=0.33\textwidth]{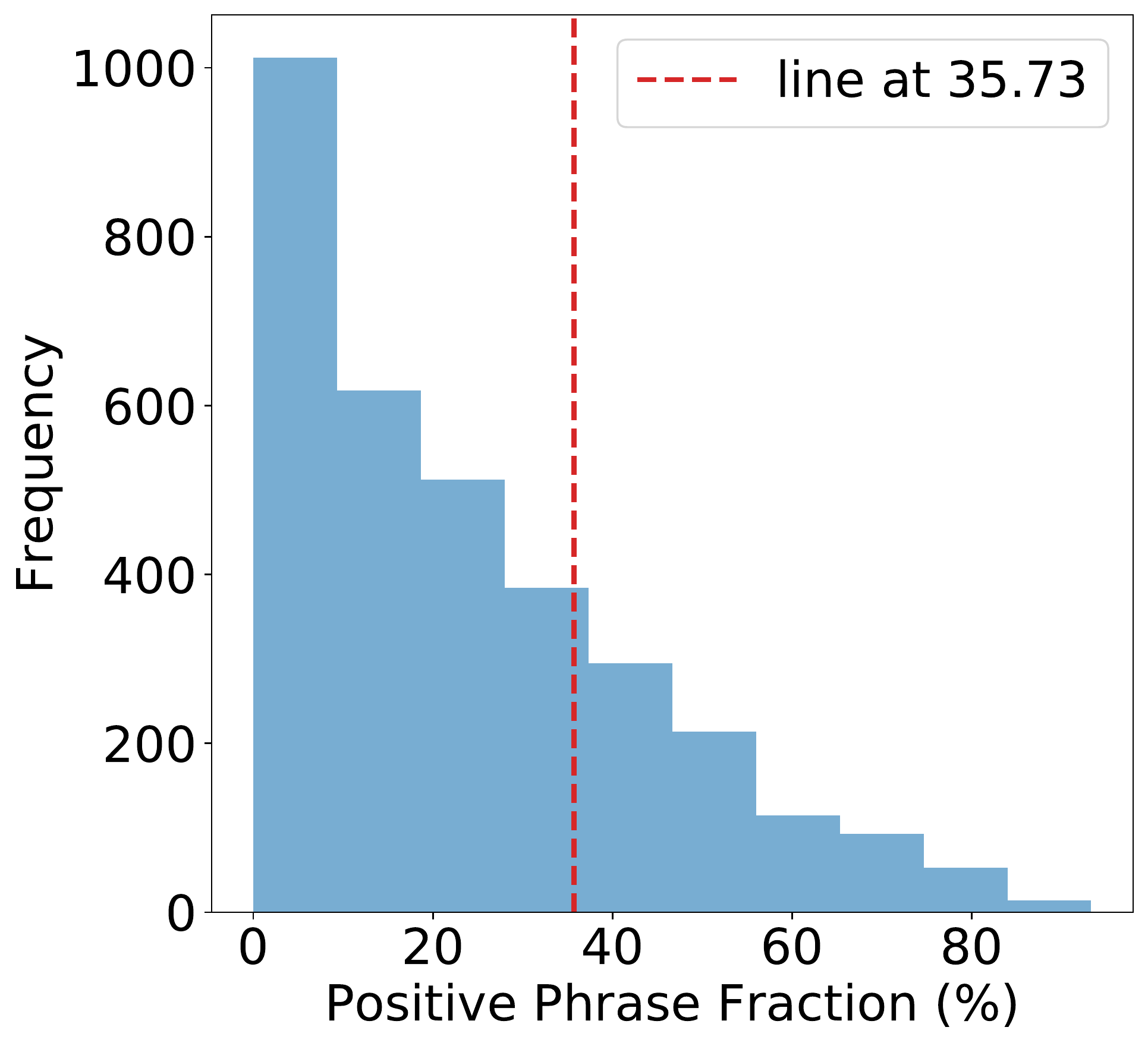}}

\subfigure[Positive Class]{
\label{Fig.Atypical.2}
\includegraphics[width=0.33\textwidth]{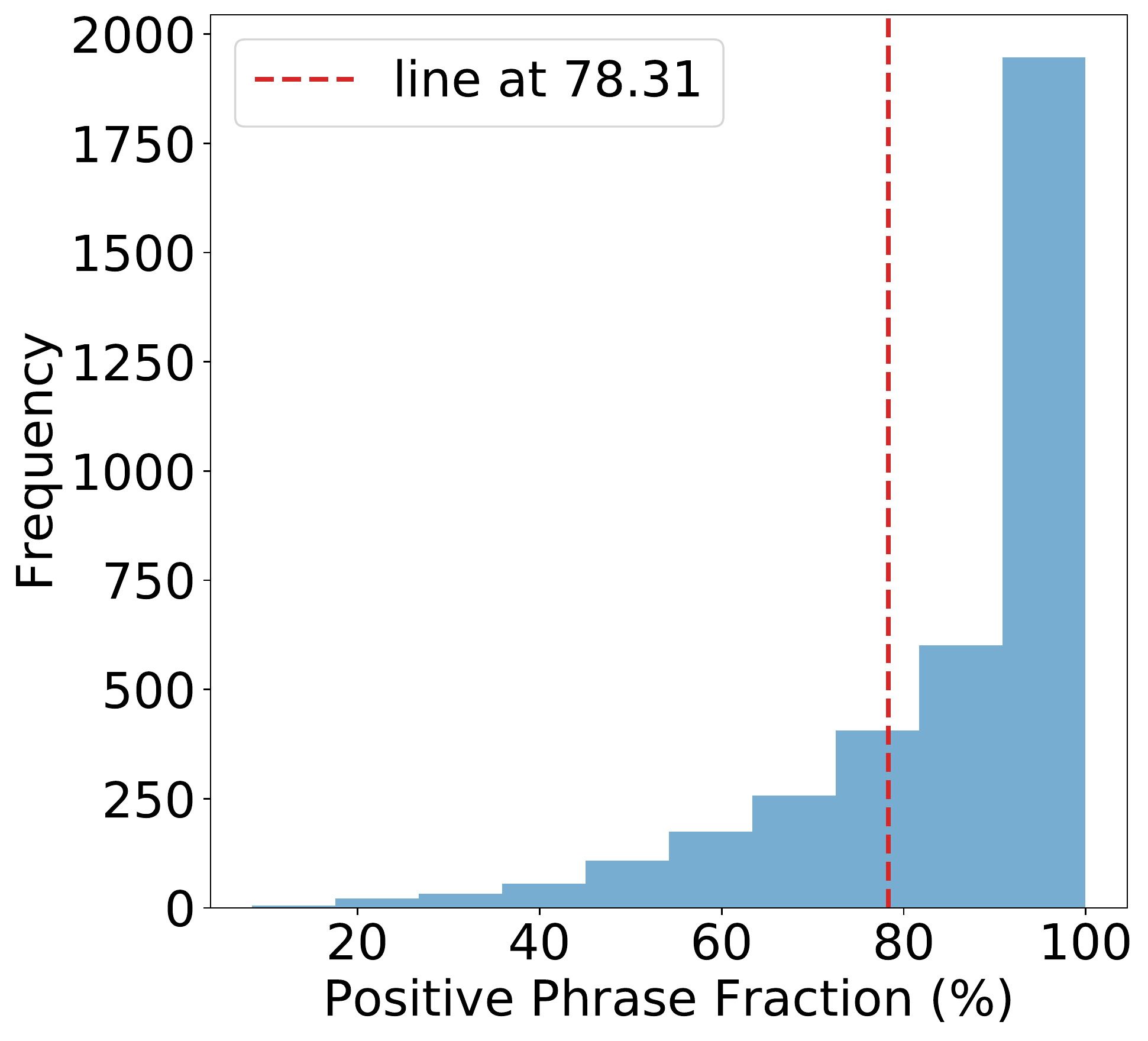}}
\caption{The distribution of positive phrase fraction on SST-2.}
\label{Fig.Atypical}
\end{figure}

We plot the distributions of positive phrase fractions for both positive instances and negative instances.
The results are shown in Figure~\ref{Fig.Atypical}. 
The results demonstrate that if we use the positive fraction to characterize the SST-2 data, then SST-2 instances of each class follow a long-tail distribution, and in the main body of our paper, we show that the top-memorized positive and negative instances likely lie in the tail end of the two distributions, judging by their average positive fraction value.

\section{Examples of Memorization Attribution}
\label{appendix:attribution}
Some examples of Memorization Attribution on Yahoo! Answers are shown in Table~\ref{table.vis.yahoo.attr}. 
In particular, without any cherry-picking, we select those instances with the highest memorization scores to present.
We can observe that on Yahoo! Answers, for most cases, the model tends to memorize those atypical parts of an instance. For example, the model needs to memorize the word ``business" that shows up in an instance labeled as ``Health" and the word ``sports" in the ``Education \& Reference" instance. 
However, one might wonder why words like ``football" and ``field" received high memorization scores for the example in ``Sports".
Although we are not certain, we hypothesize that this might be due to the fact that the span ``football field" is  atypical for the ``Sports'' category, because we find that this span shows up in only 2 instances out of 1000 ``Sports" instance in our training set.

\begin{table*}
\centering
\resizebox{\textwidth}{!}{
\begin{small}
\begin{tabular}{ll}
\toprule
\textbf{Content} &\textbf{Label} \\
\midrule
\begin{CJK*}{UTF8}{gbsn}
{\setlength{\fboxsep}{0pt}\colorbox{white!0}{\parbox{0.8\textwidth}{
\colorbox{blue!1.0860962}{\strut why} \colorbox{blue!1.1465623}{\strut are} \colorbox{red!13.290523}{\strut americans} \colorbox{red!0.5923379}{\strut .} \colorbox{red!0.3585841}{\strut .} \colorbox{red!0.13564578}{\strut .} \colorbox{blue!0.2654652}{\strut ?} \colorbox{red!0.33691382}{\strut ;} \colorbox{blue!1.3224796}{\strut why} \colorbox{blue!1.525895}{\strut are} \colorbox{red!9.366374}{\strut americans} \colorbox{red!2.7671862}{\strut so} \colorbox{blue!0.14482135}{\strut obsesed} \colorbox{blue!1.5556873}{\strut with} \colorbox{blue!12.183321}{\strut saying} \colorbox{red!0.08579347}{\strut "} \colorbox{blue!11.887416}{\strut god} \colorbox{blue!2.4510968}{\strut bless} \colorbox{red!47.380985}{\strut america} \colorbox{blue!0.07895005}{\strut "} \colorbox{red!0.11834274}{\strut .} \colorbox{red!0.20863649}{\strut i} \colorbox{blue!0.18778412}{\strut mean} \colorbox{blue!0.20577356}{\strut there} \colorbox{blue!0.19367726}{\strut is} \colorbox{red!0.2542028}{\strut no} \colorbox{blue!0.6925954}{\strut other} \colorbox{red!9.608899}{\strut country} \colorbox{blue!0.6069354}{\strut in} \colorbox{blue!0.18588664}{\strut the} \colorbox{blue!3.0798364}{\strut world} \colorbox{red!0.722552}{\strut that} \colorbox{red!0.45905727}{\strut says} \colorbox{red!0.9702893}{\strut that} \colorbox{red!0.51257485}{\strut .} \colorbox{blue!0.24146777}{\strut why} \colorbox{blue!0.25469032}{\strut must} \colorbox{blue!3.6770687}{\strut god} \colorbox{blue!2.3493397}{\strut bless} \colorbox{red!1.5920208}{\strut them} \colorbox{blue!0.075253606}{\strut when} \colorbox{red!0.6347652}{\strut they} \colorbox{red!0.45722774}{\strut have} \colorbox{red!0.777109}{\strut been} \colorbox{red!4.519401}{\strut involved} \colorbox{red!2.3250716}{\strut in} \colorbox{red!0.02456767}{\strut nearly} \colorbox{blue!0.2897979}{\strut every} \colorbox{red!75.60537}{\strut war} \colorbox{red!0.659956}{\strut to} \colorbox{red!1.2955787}{\strut date} \colorbox{red!0.7490968}{\strut .} \colorbox{blue!0.11388558}{\strut i} \colorbox{red!0.464753}{\strut '} \colorbox{red!0.28244275}{\strut m} \colorbox{red!0.28108516}{\strut not} \colorbox{blue!0.050590996}{\strut trying} \colorbox{red!0.024969421}{\strut to} \colorbox{blue!0.6090161}{\strut insult} \colorbox{red!1.2236214}{\strut them} \colorbox{blue!0.008803379}{\strut or} \colorbox{red!0.27593973}{\strut anything} \colorbox{red!0.09179174}{\strut but} \colorbox{blue!0.21976219}{\strut why} \colorbox{red!0.015724258}{\strut do} \colorbox{red!0.9065815}{\strut they} \colorbox{red!0.5303135}{\strut do} \colorbox{red!0.60973656}{\strut it} \colorbox{red!0.07267778}{\strut ?} \colorbox{red!0.49593133}{\strut ;} \colorbox{red!0.31099135}{\strut we} \colorbox{blue!0.16448024}{\strut are} \colorbox{red!0.009459437}{\strut a} \colorbox{red!11.964033}{\strut nation} \colorbox{red!1.7465193}{\strut under} \colorbox{blue!15.303607}{\strut god} \colorbox{red!0.5119772}{\strut ,} \colorbox{red!0.4186698}{\strut we} \colorbox{red!0.33056337}{\strut was} \colorbox{red!0.3751858}{\strut founded} \colorbox{red!0.085723445}{\strut from} \colorbox{red!0.43239248}{\strut it} \colorbox{red!0.31582215}{\strut .} \colorbox{red!0.29209784}{\strut .} \colorbox{red!0.09567371}{\strut .}  \colorbox{red!0.5673065}{\strut it} \colorbox{red!0.30468804}{\strut is} \colorbox{red!0}{\strut our} \colorbox{red!0.2543685}{\strut of} \colorbox{blue!0.08423959}{\strut respect} \colorbox{red!0.28584975}{\strut of} \colorbox{red!1.5945107}{\strut leader} \colorbox{red!0.39959735}{\strut of} \colorbox{red!0.13648187}{\strut our} \colorbox{red!8.0139065}{\strut country} \colorbox{red!0.018635714}{\strut before} \colorbox{red!0.33902463}{\strut us} \colorbox{red!0.45462582}{\strut ,} \colorbox{blue!0.04545475}{\strut and} \colorbox{red!0.0018709096}{\strut the} \colorbox{blue!1.0531263}{\strut great} \colorbox{red!0.841208}{\strut leader} \colorbox{red!0.26677582}{\strut in} \colorbox{blue!7.6506467}{\strut heaven} \colorbox{blue!3.142228}{\strut god} \colorbox{red!0.32801363}{\strut .} \colorbox{red!0.21772496}{\strut .} 
}}}
\end{CJK*} & Society \& Culture \\ 
\midrule
\begin{CJK*}{UTF8}{gbsn}
{\setlength{\fboxsep}{0pt}\colorbox{white!0}{\parbox{0.8\textwidth}{
\colorbox{blue!2.1896005}{\strut is} \colorbox{red!1.543739}{\strut it} \colorbox{red!2.7891607}{\strut possible} \colorbox{red!4.1683335}{\strut for} \colorbox{red!3.6649895}{\strut seven} \colorbox{red!1.9163711}{\strut 375} \colorbox{red!21.36455}{\strut pound} \colorbox{red!25.22485}{\strut men} \colorbox{red!1.9712368}{\strut to} \colorbox{red!1.402833}{\strut stand} \colorbox{red!0.8274691}{\strut on} \colorbox{blue!0.45473227}{\strut top} \colorbox{red!0.7848413}{\strut of} \colorbox{red!0.37247333}{\strut a} \colorbox{red!0.44733897}{\strut bus} \colorbox{red!0.663997}{\strut and} \colorbox{red!12.43132}{\strut pee} \colorbox{red!4.2702055}{\strut while} \colorbox{red!0.38512906}{\strut it} \colorbox{red!16.709694}{\strut races} \colorbox{red!0.5807025}{\strut down} \colorbox{red!3.0815392}{\strut the} \colorbox{red!3.809683}{\strut hi} \colorbox{red!2.58003}{\strut -} \colorbox{red!4.772446}{\strut way} \colorbox{red!1.0318472}{\strut ?} \colorbox{red!0.6302896}{\strut ;} \colorbox{red!3.4689646}{\strut they} \colorbox{blue!0.14194205}{\strut would} \colorbox{red!2.1183014}{\strut be} \colorbox{red!12.056758}{\strut belted} \colorbox{red!2.1457727}{\strut in} \colorbox{red!1.0404257}{\strut of} \colorbox{red!1.1133604}{\strut course} \colorbox{red!1.8090152}{\strut for} \colorbox{red!11.315616}{\strut safety} \colorbox{red!1.9472989}{\strut reasons} \colorbox{blue!1.553577}{\strut ,} \colorbox{red!2.210458}{\strut so} \colorbox{red!0.7422025}{\strut the} \colorbox{red!3.1795487}{\strut formula} \colorbox{red!0.72100663}{\strut is} \colorbox{red!0.9864082}{\strut seven} \colorbox{red!0.32588238}{\strut 375} \colorbox{red!6.7020884}{\strut pound} \colorbox{red!5.194278}{\strut men} \colorbox{red!0.10354689}{\strut ,} \colorbox{red!3.519775}{\strut seat} \colorbox{red!0.57261115}{\strut -} \colorbox{red!3.0417693}{\strut belted} \colorbox{red!0.49213082}{\strut on} \colorbox{red!0.2161832}{\strut top} \colorbox{red!0.5092522}{\strut of} \colorbox{blue!0.49795088}{\strut a} \colorbox{blue!1.3792019}{\strut bus} \colorbox{red!0.22034256}{\strut ,} \colorbox{red!9.52999}{\strut peeing} \colorbox{red!1.1597899}{\strut at} \colorbox{red!1.423256}{\strut 75} \colorbox{blue!2.99156}{\strut miles} \colorbox{red!0.45191142}{\strut per} \colorbox{red!0.52449274}{\strut hour} \colorbox{red!0.21851125}{\strut ,} \colorbox{blue!0.010387515}{\strut into} \colorbox{red!0.5146051}{\strut a} \colorbox{red!0.7636938}{\strut head} \colorbox{red!0.5747576}{\strut -} \colorbox{blue!3.6000462}{\strut wind} \colorbox{red!0.4911219}{\strut of} \colorbox{red!0.91742724}{\strut 10} \colorbox{red!1.0202985}{\strut mph} \colorbox{red!0.3738925}{\strut ,} \colorbox{red!0.54571533}{\strut at} \colorbox{red!0.55409986}{\strut a} \colorbox{red!1.0445201}{\strut 30} \colorbox{red!0.25628367}{\strut degree} \colorbox{red!0.3907782}{\strut angle} \colorbox{red!1.4782333}{\strut ,} \colorbox{blue!0.010278304}{\strut what} \colorbox{blue!0.23906398}{\strut is} \colorbox{blue!0.110508956}{\strut the} \colorbox{red!3.7602634}{\strut end} \colorbox{blue!2.5865586}{\strut velocity} \colorbox{red!0.3852071}{\strut of} \colorbox{red!0.95210683}{\strut the} \colorbox{red!12.221786}{\strut pee} \colorbox{red!0.54333794}{\strut ?} \colorbox{red!0.7223919}{\strut ?} \colorbox{red!1.1482066}{\strut ;} \colorbox{red!0.32861504}{\strut first} \colorbox{red!0.43650642}{\strut of} \colorbox{red!0.69243807}{\strut all} \colorbox{red!0.121158235}{\strut it} \colorbox{red!5.5267286}{\strut wont} \colorbox{red!1.7961512}{\strut look} \colorbox{red!0.56394243}{\strut too} \colorbox{red!3.8616874}{\strut good} \colorbox{red!0.95044273}{\strut .} \colorbox{red!0.5836174}{\strut .} \colorbox{red!1.2301043}{\strut .} \colorbox{red!3.3472474}{\strut thats} \colorbox{red!0.70275766}{\strut a} \colorbox{red!2.162697}{\strut lot} \colorbox{red!1.1891309}{\strut of} \colorbox{red!11.924183}{\strut pee} \colorbox{red!1.1971079}{\strut !} \colorbox{red!1.3304622}{\strut !} \colorbox{red!1.3006142}{\strut !} \colorbox{blue!0.0613938}{\strut next} \colorbox{red!0.62890023}{\strut ,} \colorbox{red!3.3922434}{\strut they} \colorbox{blue!0.533835}{\strut must} \colorbox{red!1.2889097}{\strut have} \colorbox{red!0.9224657}{\strut on} \colorbox{red!6.252714}{\strut water} \colorbox{red!5.5915008}{\strut proff} \colorbox{red!1.425863}{\strut clothing} \colorbox{red!1.6240183}{\strut ,} \colorbox{blue!0.32671723}{\strut it} \colorbox{red!2.160842}{\strut will} 
}}}
\end{CJK*} & Science \& Mathematics \\ 
\midrule
\begin{CJK*}{UTF8}{gbsn}
{\setlength{\fboxsep}{0pt}\colorbox{white!0}{\parbox{0.8\textwidth}{
\colorbox{red!1.9198865}{\strut what} \colorbox{red!3.766017}{\strut would} \colorbox{red!5.4044}{\strut you} \colorbox{red!4.483419}{\strut do} \colorbox{red!2.5183997}{\strut ?} \colorbox{red!1.7686688}{\strut ;} \colorbox{red!1.7079257}{\strut i} \colorbox{red!1.3555788}{\strut have} \colorbox{red!1.2416714}{\strut an} \colorbox{red!4.880533}{\strut opportunity} \colorbox{red!3.54295}{\strut to} \colorbox{red!3.4852943}{\strut take} \colorbox{red!3.2267072}{\strut over} \colorbox{red!3.7418694}{\strut a} \colorbox{red!45.142654}{\strut business} \colorbox{red!4.1736712}{\strut in} \colorbox{red!5.996068}{\strut the} \colorbox{blue!21.57602}{\strut womans} \colorbox{blue!60.31795}{\strut health} \colorbox{red!6.383597}{\strut field} \colorbox{red!1.7944418}{\strut ,} \colorbox{red!0.23686036}{\strut with} \colorbox{red!1.5650597}{\strut a} \colorbox{red!2.2827532}{\strut solid} \colorbox{red!13.825377}{\strut cash} \colorbox{red!2.700628}{\strut flow} \colorbox{blue!0.048412405}{\strut but} \colorbox{red!0.94204664}{\strut part} \colorbox{red!1.6948024}{\strut of} \colorbox{red!1.9773226}{\strut the} \colorbox{red!13.399037}{\strut deal} \colorbox{blue!0.72201455}{\strut means} \colorbox{red!1.3011428}{\strut i} \colorbox{red!2.2553217}{\strut must} \colorbox{red!1.1859125}{\strut take} \colorbox{red!1.5199351}{\strut over} \colorbox{red!1.3493216}{\strut an} \colorbox{red!1.8017695}{\strut additional} \colorbox{red!4.7695985}{\strut location} \colorbox{red!0.67586064}{\strut that} \colorbox{red!0.35567185}{\strut has} \colorbox{red!0.64243126}{\strut a} \colorbox{red!0.1845583}{\strut negative} \colorbox{red!4.0082374}{\strut cash} \colorbox{red!1.0575813}{\strut flow} \colorbox{red!1.2853451}{\strut .} \colorbox{red!0.39583576}{\strut i} \colorbox{red!0.5333033}{\strut have} \colorbox{red!0.38987225}{\strut enough} \colorbox{red!1.1653149}{\strut money} \colorbox{red!0.383001}{\strut to} \colorbox{red!1.393822}{\strut pay} \colorbox{red!0.4238165}{\strut for} \colorbox{red!1.3741286}{\strut the} \colorbox{red!14.647591}{\strut business} \colorbox{red!0.23390791}{\strut and} \colorbox{red!0.3267578}{\strut a} \colorbox{red!0.08201809}{\strut little} \colorbox{red!0.23571005}{\strut left} \colorbox{red!0.33277065}{\strut over} \colorbox{red!0.30667755}{\strut to} \colorbox{red!0.68718415}{\strut satisfy} \colorbox{red!0.66773105}{\strut a} \colorbox{red!1.6918775}{\strut shortfall} \colorbox{red!0.84401554}{\strut in} \colorbox{red!4.318705}{\strut operating} \colorbox{red!6.5224366}{\strut cash} \colorbox{red!1.8090191}{\strut flow} \colorbox{red!0.7797017}{\strut of} \colorbox{red!0.60485786}{\strut just} \colorbox{red!1.0848376}{\strut the} \colorbox{red!0.39360622}{\strut one} \colorbox{red!1.1865925}{\strut .} \colorbox{red!0.6296266}{\strut i} \colorbox{red!0.9095354}{\strut did} \colorbox{red!0.4773031}{\strut not} \colorbox{red!1.5533444}{\strut factor} \colorbox{red!0.7383998}{\strut anything} \colorbox{red!0.7854147}{\strut in} \colorbox{red!0.6946005}{\strut for} \colorbox{red!1.2880194}{\strut the} \colorbox{red!0.9819226}{\strut second} \colorbox{red!2.2087858}{\strut location} \colorbox{red!0.33518428}{\strut with} \colorbox{red!0.38555586}{\strut a} \colorbox{red!0.37191984}{\strut negative} \colorbox{red!2.982171}{\strut operating} \colorbox{red!3.5165064}{\strut cash} \colorbox{red!1.3948355}{\strut flow} \colorbox{red!1.2961379}{\strut .} \colorbox{red!0.7138066}{\strut the} \colorbox{red!2.4570837}{\strut crunch} \colorbox{red!0.69450533}{\strut is} \colorbox{red!0.572929}{\strut that} \colorbox{red!0.19541681}{\strut i} \colorbox{red!0.65584147}{\strut can} \colorbox{red!0.52102757}{\strut not} \colorbox{red!0.6879653}{\strut have} \colorbox{red!0.8984927}{\strut one} \colorbox{red!0.3325498}{\strut without} \colorbox{red!1.0574671}{\strut the} \colorbox{red!0.9430697}{\strut other} \colorbox{red!1.27503}{\strut .} \colorbox{red!0.608378}{\strut the} \colorbox{red!0.81096524}{\strut important} \colorbox{red!0.58537734}{\strut thing} \colorbox{red!0.48102066}{\strut is} \colorbox{red!0.6628109}{\strut to} \colorbox{red!0.8451052}{\strut know} \colorbox{red!0.56192493}{\strut that} \colorbox{blue!0.65124774}{\strut i} \colorbox{red!0.3113435}{\strut am} \colorbox{red!1.0696579}{\strut only} \colorbox{red!1.1951087}{\strut short} \colorbox{red!4.466255}{\strut operating} \colorbox{red!2.263183}{\strut capitol} \colorbox{red!2.0033202}{\strut for} \colorbox{red!1.216756}{\strut one} \colorbox{red!5.06013}{\strut location} \colorbox{red!0.73087937}{\strut .} \colorbox{red!0.51403004}{\strut .} \colorbox{red!0.46539208}{\strut .} \colorbox{red!0.27175722}{\strut .} \colorbox{red!2.2122447}{\strut should} 
}}}
\end{CJK*} & Health \\ 
\midrule
\begin{CJK*}{UTF8}{gbsn}
{\setlength{\fboxsep}{0pt}\colorbox{white!0}{\parbox{0.8\textwidth}{
\colorbox{red!0.31430823}{\strut my} \colorbox{red!1.3844963}{\strut hs} \colorbox{blue!2.1827633}{\strut son} \colorbox{red!19.286385}{\strut plays} \colorbox{blue!8.901905}{\strut two} \colorbox{red!4.3314443}{\strut hs} \colorbox{red!71.71763}{\strut sports} \colorbox{blue!1.0375654}{\strut -} \colorbox{red!2.283319}{\strut hardly} \colorbox{red!1.4738784}{\strut find} \colorbox{red!0.10713821}{\strut time} \colorbox{red!3.9663289}{\strut for} \colorbox{blue!0.065747984}{\strut h} \colorbox{red!1.6956501}{\strut /} \colorbox{red!1.6115034}{\strut w} \colorbox{blue!0.017851977}{\strut -} \colorbox{blue!0.49599296}{\strut i} \colorbox{blue!0.20842496}{\strut want} \colorbox{blue!0.60669565}{\strut to} \colorbox{red!0.39790842}{\strut send} \colorbox{red!1.9994935}{\strut him} \colorbox{red!2.230575}{\strut to} \colorbox{blue!0.44775018}{\strut prep} \colorbox{blue!6.9152036}{\strut school} \colorbox{blue!0.7880608}{\strut to} \colorbox{blue!0.84458685}{\strut imprv} \colorbox{blue!0.10877475}{\strut his} \colorbox{blue!10.888408}{\strut grades} \colorbox{red!0.7015338}{\strut ;} \colorbox{red!0.16527377}{\strut i} \colorbox{red!0.079473294}{\strut want} \colorbox{red!0.31957594}{\strut him} \colorbox{red!0.15835811}{\strut to} \colorbox{red!0.100432485}{\strut have} \colorbox{red!0.22976443}{\strut a} \colorbox{red!0.1551997}{\strut high} \colorbox{blue!11.25381}{\strut sat} \colorbox{red!0.7246978}{\strut /} \colorbox{blue!5.1609745}{\strut act} \colorbox{red!0.41408265}{\strut as} \colorbox{red!0.27982143}{\strut well} \colorbox{red!0.25382388}{\strut high} \colorbox{blue!1.561982}{\strut gpa} \colorbox{blue!0.1717259}{\strut to} \colorbox{red!0.9273658}{\strut go} \colorbox{red!1.0811378}{\strut in} \colorbox{red!1.2725104}{\strut to} \colorbox{blue!7.186629}{\strut college} \colorbox{blue!0.3182647}{\strut .} \colorbox{red!0.22476737}{\strut i} \colorbox{red!0.8841041}{\strut hear} \colorbox{red!0.0077390615}{\strut that} \colorbox{red!0.83037436}{\strut prep} \colorbox{blue!2.7153401}{\strut boarding} \colorbox{blue!11.808243}{\strut schools} \colorbox{blue!0.012039072}{\strut can} \colorbox{blue!0.07222896}{\strut be} \colorbox{blue!1.8384464}{\strut expensive} \colorbox{blue!0.060913973}{\strut .} \colorbox{red!0.009844118}{\strut i} \colorbox{red!0.0763593}{\strut need} \colorbox{blue!0.1879956}{\strut help} \colorbox{red!0.36121088}{\strut !} \colorbox{red!0.39305738}{\strut ;} \colorbox{red!0.266919}{\strut i} \colorbox{red!0.36006948}{\strut know} \colorbox{blue!0.3973331}{\strut this} \colorbox{red!0.06658695}{\strut will} \colorbox{blue!0.08646586}{\strut sound} \colorbox{red!0.87406105}{\strut cold} \colorbox{red!0.102363214}{\strut and} \colorbox{red!0.5302209}{\strut uncaring} \colorbox{red!0.47777444}{\strut but} \colorbox{red!0.16170248}{\strut really} \colorbox{blue!0.21283291}{\strut it} \colorbox{red!0.12536639}{\strut '} \colorbox{red!0.036789663}{\strut s} \colorbox{red!0.11945438}{\strut not} \colorbox{blue!0.35363963}{\strut .}  \colorbox{red!1.8213835}{\strut if} \colorbox{red!0.64494896}{\strut he} \colorbox{red!0.055237707}{\strut '} \colorbox{red!0.20906524}{\strut s} \colorbox{red!0.0057072053}{\strut having} \colorbox{red!1.734929}{\strut probs} \colorbox{red!0.20748228}{\strut with} \colorbox{red!24.126774}{\strut sports} \colorbox{red!0.12650162}{\strut and} \colorbox{red!0.63887674}{\strut keeping} \colorbox{blue!5.358803}{\strut grades} \colorbox{red!0.4633775}{\strut up} \colorbox{red!0.2092347}{\strut .} \colorbox{blue!0.09352193}{\strut .} \colorbox{red!0.16712053}{\strut .} \colorbox{red!0.2160177}{\strut take} \colorbox{red!0.030709255}{\strut away} \colorbox{blue!0.20115939}{\strut the} \colorbox{red!29.318487}{\strut sports} \colorbox{blue!0.33917987}{\strut privileges} \colorbox{red!0.47021517}{\strut .} \colorbox{blue!16.679653}{\strut school} \colorbox{blue!0.9509845}{\strut work} \colorbox{blue!0.3486641}{\strut should} \colorbox{blue!0.04819624}{\strut be} \colorbox{red!0.5207729}{\strut his} \colorbox{red!0.28849676}{\strut main} \colorbox{red!0.016841905}{\strut focus} \colorbox{red!0.20407338}{\strut ,} \colorbox{blue!0.013400527}{\strut then} \colorbox{red!10.934539}{\strut sports} \colorbox{red!0.03711953}{\strut .}  \colorbox{red!1.8855866}{\strut my} \colorbox{blue!0.24308388}{\strut son} \colorbox{blue!0.57488894}{\strut is} \colorbox{blue!0.29836673}{\strut in} \colorbox{blue!0.9289006}{\strut a} 
}}}
\end{CJK*} & Education \& Reference \\
\midrule
\begin{CJK*}{UTF8}{gbsn}
{\setlength{\fboxsep}{0pt}\colorbox{white!0}{\parbox{0.8\textwidth}{
\colorbox{red!2.0028222}{\strut out} \colorbox{red!3.1217518}{\strut of} \colorbox{red!2.2716393}{\strut all} \colorbox{red!5.756519}{\strut the} \colorbox{red!41.326553}{\strut schools} \colorbox{red!1.8181795}{\strut in} \colorbox{red!0.42149538}{\strut nigeria} \colorbox{blue!0.9125999}{\strut that} \colorbox{red!1.4118205}{\strut have} \colorbox{blue!16.657906}{\strut computers} \colorbox{red!4.6038156}{\strut ,} \colorbox{red!4.803349}{\strut how} \colorbox{red!3.5436778}{\strut many} \colorbox{red!9.681702}{\strut have} \colorbox{red!18.36809}{\strut internet} \colorbox{red!11.650192}{\strut access} \colorbox{red!3.3076742}{\strut ?} \colorbox{red!5.46506}{\strut ;} \colorbox{red!2.8188312}{\strut i} \colorbox{red!3.2945235}{\strut '} \colorbox{red!1.6397065}{\strut m} \colorbox{red!3.310961}{\strut looking} \colorbox{red!2.1126838}{\strut into} \colorbox{blue!0.948161}{\strut some} \colorbox{red!41.339664}{\strut overseas} \colorbox{red!55.3232}{\strut development} \colorbox{red!20.203287}{\strut ideas} \colorbox{red!1.3972188}{\strut .} \colorbox{red!1.123057}{\strut do} \colorbox{red!1.5764939}{\strut you} \colorbox{red!1.9652684}{\strut know} \colorbox{red!3.0134828}{\strut roughly} \colorbox{red!2.6009161}{\strut what} \colorbox{red!1.6294006}{\strut percentage} \colorbox{red!1.4197737}{\strut have} \colorbox{red!3.8401253}{\strut internet} \colorbox{red!3.382302}{\strut access} \colorbox{red!1.7148007}{\strut (} \colorbox{red!1.2782382}{\strut most} \colorbox{red!1.4055202}{\strut or} \colorbox{red!0.8874477}{\strut just} \colorbox{red!0.9647379}{\strut a} \colorbox{red!0.99053335}{\strut few} \colorbox{red!2.2987666}{\strut )} \colorbox{red!1.0645001}{\strut ?} \colorbox{red!2.5896769}{\strut ;} \colorbox{blue!2.539664}{\strut my} \colorbox{red!2.6306393}{\strut school} \colorbox{blue!5.2262697}{\strut in} \colorbox{red!1.1513875}{\strut nigeria} \colorbox{red!0.26102903}{\strut had} \colorbox{red!11.069925}{\strut internet} \colorbox{red!3.9760048}{\strut service} \colorbox{red!0.23460414}{\strut ,} \colorbox{red!4.41045}{\strut it} \colorbox{red!3.0634234}{\strut is} \colorbox{red!2.2801647}{\strut the} \colorbox{red!1.9606547}{\strut best} \colorbox{red!25.152912}{\strut school} \colorbox{red!2.2291534}{\strut i} \colorbox{red!1.630301}{\strut have} \colorbox{blue!1.458287}{\strut seen} \colorbox{red!0.7158}{\strut till} \colorbox{red!0.3855864}{\strut today} \colorbox{red!1.2493019}{\strut .} \colorbox{red!1.840749}{\strut .} \colorbox{red!0.78602487}{\strut .} 
}}}
\end{CJK*} & Computers \& Internet \\
\midrule
\begin{CJK*}{UTF8}{gbsn}
{\setlength{\fboxsep}{0pt}\colorbox{white!0}{\parbox{0.8\textwidth}{
\colorbox{red!2.758984}{\strut where} \colorbox{red!2.9519844}{\strut does} \colorbox{red!0.76688707}{\strut the} \colorbox{blue!4.1057835}{\strut term} \colorbox{red!8.118242}{\strut grid} \colorbox{red!12.194696}{\strut iron} \colorbox{red!5.2356186}{\strut originate} \colorbox{red!1.6588427}{\strut and} \colorbox{red!1.4304872}{\strut how} \colorbox{red!1.5520799}{\strut did} \colorbox{red!2.2854798}{\strut it} \colorbox{red!2.4171205}{\strut get} \colorbox{red!4.2731247}{\strut applied} \colorbox{red!0.9079076}{\strut to} \colorbox{red!6.3629947}{\strut a} \colorbox{red!42.85321}{\strut football} \colorbox{red!24.835474}{\strut field} \colorbox{red!2.1660135}{\strut .} \colorbox{red!1.1936957}{\strut ?} \colorbox{red!1.1778883}{\strut ;} \colorbox{red!1.2078496}{\strut what} \colorbox{red!1.3767453}{\strut is} \colorbox{red!1.16399}{\strut the} \colorbox{red!0.2485368}{\strut original} \colorbox{red!4.8654213}{\strut meaning} \colorbox{blue!0.7371815}{\strut of} \colorbox{red!11.490101}{\strut gridiron} \colorbox{red!1.0424199}{\strut .} \colorbox{red!0.6586075}{\strut who} \colorbox{red!1.8757155}{\strut applied} \colorbox{red!2.285905}{\strut it} \colorbox{red!1.4355608}{\strut to} \colorbox{red!4.2604094}{\strut a} \colorbox{red!24.407356}{\strut football} \colorbox{red!12.633708}{\strut field} \colorbox{red!1.0932597}{\strut and} \colorbox{red!0.6868579}{\strut why} \colorbox{red!0.7037872}{\strut ?} \colorbox{red!0.86770666}{\strut ;} \colorbox{red!0.76799744}{\strut hi} \colorbox{red!1.0935116}{\strut there} \colorbox{red!0.81606174}{\strut .} \colorbox{red!0.7770301}{\strut .} \colorbox{red!0.7663185}{\strut .}   \colorbox{red!1.6304482}{\strut here} \colorbox{red!1.0787543}{\strut is} \colorbox{red!0.9791744}{\strut the} \colorbox{red!1.3339211}{\strut answer} \colorbox{red!0.8396944}{\strut i} \colorbox{red!0.8914847}{\strut found} \colorbox{red!0.6069195}{\strut from} \colorbox{red!0.7671665}{\strut the} \colorbox{red!1.0675522}{\strut word} \colorbox{blue!1.7839955}{\strut detective} \colorbox{red!1.7586049}{\strut site} \colorbox{red!0.75775754}{\strut :}   \colorbox{red!1.7870016}{\strut the} \colorbox{red!0.71036154}{\strut use} \colorbox{red!0.4413459}{\strut of} \colorbox{red!0.6724581}{\strut "} \colorbox{red!7.4650817}{\strut gridiron} \colorbox{red!0.78964543}{\strut "} \colorbox{red!0.50203365}{\strut as} \colorbox{red!0.4904877}{\strut a} \colorbox{red!2.0143762}{\strut metaphor} \colorbox{red!0.73293287}{\strut for} \colorbox{red!2.7360613}{\strut the} \colorbox{red!22.600933}{\strut football} \colorbox{red!7.564591}{\strut field} \colorbox{red!0.098087706}{\strut ,} \colorbox{red!0.6501783}{\strut and} \colorbox{red!0.2756989}{\strut ,} \colorbox{red!0.442171}{\strut by} \colorbox{red!0.66364604}{\strut extension} \colorbox{red!0.4248662}{\strut ,} \colorbox{red!0.6653365}{\strut to} \colorbox{red!1.4563086}{\strut the} \colorbox{red!4.6034718}{\strut game} \colorbox{red!0.4540147}{\strut itself} \colorbox{red!0.4976471}{\strut ,} \colorbox{red!0.88102996}{\strut dates} \colorbox{red!0.5001627}{\strut back} \colorbox{red!0.5029643}{\strut to} \colorbox{red!2.10294}{\strut 1897} \colorbox{red!0.089454226}{\strut .} \colorbox{red!0.9855846}{\strut the} \colorbox{red!0.8594631}{\strut original} \colorbox{red!0.6244388}{\strut "} \colorbox{red!7.4179354}{\strut gridirons} \colorbox{red!0.8677465}{\strut "} \colorbox{red!1.0665894}{\strut were} \colorbox{red!0.83046484}{\strut just} \colorbox{red!0.82725257}{\strut that} \colorbox{red!1.508304}{\strut :} \colorbox{red!4.674805}{\strut grids} \colorbox{red!1.1069021}{\strut made} \colorbox{red!0.75722533}{\strut of} \colorbox{red!2.4410474}{\strut iron} \colorbox{red!0.9371302}{\strut ,} \colorbox{red!0.23513961}{\strut used} \colorbox{red!0.75180054}{\strut to} \colorbox{red!0.638643}{\strut cook} \colorbox{red!0.52191275}{\strut fish} \colorbox{red!0.8286178}{\strut or} \colorbox{red!0.40236387}{\strut meat} \colorbox{red!0.64070404}{\strut over} \colorbox{red!0.30931675}{\strut an} \colorbox{red!0.64109313}{\strut open} \colorbox{red!0.016540159}{\strut fire} \colorbox{red!2.0608325}{\strut .} \colorbox{red!1.8494848}{\strut early} 
}}}
\end{CJK*} & Sports \\
\midrule
\begin{CJK*}{UTF8}{gbsn}
{\setlength{\fboxsep}{0pt}\colorbox{white!0}{\parbox{0.8\textwidth}{
\colorbox{red!1.6715776}{\strut what} \colorbox{red!1.5789691}{\strut kind} \colorbox{red!2.6276712}{\strut of} \colorbox{red!53.7759}{\strut math} \colorbox{red!1.7257015}{\strut would} \colorbox{red!1.0599397}{\strut i} \colorbox{red!3.3316512}{\strut need} \colorbox{red!1.7517401}{\strut to} \colorbox{red!1.6522477}{\strut be} \colorbox{red!5.2749977}{\strut a} \colorbox{red!11.556678}{\strut real} \colorbox{red!33.276295}{\strut estate} \colorbox{red!23.900663}{\strut appraiser} \colorbox{red!1.1879544}{\strut ?} \colorbox{red!1.3231394}{\strut ;} \colorbox{red!0.9731816}{\strut what} \colorbox{red!0.9340698}{\strut kind} \colorbox{red!1.6862956}{\strut of} \colorbox{red!25.214645}{\strut math} \colorbox{red!1.4260794}{\strut would} \colorbox{red!1.1256665}{\strut i} \colorbox{red!1.6487136}{\strut need} \colorbox{red!1.410879}{\strut to} \colorbox{red!2.1773834}{\strut be} \colorbox{red!2.983682}{\strut a} \colorbox{red!7.1268363}{\strut real} \colorbox{red!20.86679}{\strut estate} \colorbox{red!16.446474}{\strut appraiser} \colorbox{red!0.8667481}{\strut ?}  \colorbox{red!3.1091905}{\strut the} \colorbox{blue!2.3049653}{\strut job} \colorbox{red!1.3935188}{\strut as} \colorbox{red!1.7039098}{\strut says} \colorbox{red!1.3712894}{\strut needs} \colorbox{red!0.53265667}{\strut strong} \colorbox{red!17.387356}{\strut math} \colorbox{red!0.4217561}{\strut skills} \colorbox{red!0.8009159}{\strut ;} \colorbox{red!7.549794}{\strut geometry} \colorbox{red!0.9974649}{\strut (} \colorbox{red!1.1624907}{\strut area} \colorbox{red!0.75355405}{\strut of} \colorbox{red!0.6301139}{\strut a} \colorbox{red!0.7037944}{\strut circle} \colorbox{red!0.47606677}{\strut ,} \colorbox{red!1.6161504}{\strut rectangle} \colorbox{red!0.5030989}{\strut ,} \colorbox{red!0.9996788}{\strut triangle} \colorbox{red!0.6085152}{\strut ,} \colorbox{red!1.0266027}{\strut volume} \colorbox{red!0.70976245}{\strut of} \colorbox{red!0.57082856}{\strut a} \colorbox{red!1.4198734}{\strut rectangle} \colorbox{red!0.71506274}{\strut ,} \colorbox{red!0.5703212}{\strut etc} \colorbox{red!0.34737515}{\strut .} \colorbox{red!0.5365948}{\strut .} \colorbox{red!0.58130777}{\strut .} \colorbox{red!0.7713211}{\strut )} \colorbox{red!1.3959788}{\strut plus} \colorbox{red!2.7999146}{\strut percentages} \colorbox{red!0.84782326}{\strut ,} \colorbox{red!1.4723878}{\strut percentage} \colorbox{red!0.89364}{\strut of} \colorbox{red!0.51312345}{\strut change} \colorbox{red!1.2103227}{\strut .} \colorbox{red!0.5718677}{\strut some} \colorbox{red!1.9459268}{\strut minor} \colorbox{red!8.5555525}{\strut algebra} \colorbox{red!0.7451909}{\strut to} \colorbox{red!0.73334754}{\strut find} \colorbox{red!0.7206876}{\strut the} \colorbox{red!1.9187835}{\strut unknown} \colorbox{red!2.979935}{\strut vairable} \colorbox{red!0.9707206}{\strut in} \colorbox{red!1.3328322}{\strut the} \colorbox{red!2.5507035}{\strut percentage} \colorbox{red!4.427147}{\strut calcs} \colorbox{red!1.0224245}{\strut .} \colorbox{red!0.92904234}{\strut that} \colorbox{red!0.5103245}{\strut '} \colorbox{red!1.171384}{\strut s} \colorbox{red!0.8382471}{\strut about} \colorbox{red!1.8363366}{\strut it} \colorbox{red!0.56849986}{\strut .} 
}}}
\end{CJK*} & Business \& Finance \\
\midrule
\begin{CJK*}{UTF8}{gbsn}
{\setlength{\fboxsep}{0pt}\colorbox{white!0}{\parbox{0.8\textwidth}{
\colorbox{red!3.362549}{\strut does} \colorbox{red!1.9923449}{\strut anyone} \colorbox{red!10.99107}{\strut know} \colorbox{red!8.683733}{\strut any} \colorbox{red!22.221321}{\strut electro} \colorbox{red!18.900867}{\strut bands} \colorbox{red!1.1659918}{\strut ?} \colorbox{red!1.1513759}{\strut ;} \colorbox{red!2.6208808}{\strut does} \colorbox{red!2.7264428}{\strut anyone} \colorbox{red!4.0057216}{\strut know} \colorbox{red!7.2423234}{\strut of} \colorbox{red!5.965675}{\strut any} \colorbox{red!2.33885}{\strut good} \colorbox{red!23.437363}{\strut electro} \colorbox{red!22.877174}{\strut bands} \colorbox{red!2.67021}{\strut suck} \colorbox{red!4.6433187}{\strut as} \colorbox{red!38.375477}{\strut metric} \colorbox{red!1.1944869}{\strut and} \colorbox{red!44.983692}{\strut robots} \colorbox{red!0.14358282}{\strut in} \colorbox{blue!8.338192}{\strut disguise} \colorbox{red!0.9485154}{\strut ?} \colorbox{red!1.6753743}{\strut ;} \colorbox{red!3.1629205}{\strut hmmm} \colorbox{red!0.8752336}{\strut .} \colorbox{red!0.9623552}{\strut .} \colorbox{red!1.4313495}{\strut .} \colorbox{red!2.1676092}{\strut how} \colorbox{red!1.8865088}{\strut about} \colorbox{red!2.161093}{\strut :}  \colorbox{red!7.492121}{\strut particle}  \colorbox{red!6.6704354}{\strut lotus}  \colorbox{red!6.716273}{\strut pnuma} \colorbox{blue!8.13295}{\strut trio}  \colorbox{red!5.3660917}{\strut soulive}  \colorbox{red!1.2922385}{\strut brother} \colorbox{red!0.55220866}{\strut '} \colorbox{blue!0.29670453}{\strut s} \colorbox{blue!1.4729015}{\strut past}  \colorbox{red!2.2574892}{\strut look} \colorbox{red!3.7840042}{\strut for} \colorbox{red!4.2770243}{\strut these} \colorbox{red!1.38796}{\strut bands} \colorbox{red!0.31716093}{\strut and} \colorbox{red!0.020710971}{\strut lots} \colorbox{red!1.0550348}{\strut of} \colorbox{red!1.5410727}{\strut others} \colorbox{red!2.529775}{\strut at} \colorbox{red!2.5346587}{\strut :}  \colorbox{red!3.352334}{\strut http} \colorbox{red!1.1017323}{\strut :} \colorbox{red!0.5746026}{\strut /} \colorbox{red!0.72764426}{\strut /} \colorbox{red!0.90611327}{\strut www} \colorbox{blue!0.47752774}{\strut .} \colorbox{red!5.788042}{\strut archive} \colorbox{red!0.48838216}{\strut .} \colorbox{red!1.5057255}{\strut org} \colorbox{red!0.36568832}{\strut /} \colorbox{red!2.6741161}{\strut details} \colorbox{red!0.7377096}{\strut /} \colorbox{red!12.971097}{\strut etree} 
}}}
\end{CJK*} & Entertainment \& Music \\
\midrule
\begin{CJK*}{UTF8}{gbsn}
{\setlength{\fboxsep}{0pt}\colorbox{white!0}{\parbox{0.8\textwidth}{
\colorbox{blue!1.3987659}{\strut how} \colorbox{blue!0.6486729}{\strut can} \colorbox{blue!0.29634744}{\strut make} \colorbox{red!0.0761432}{\strut a} \colorbox{blue!3.0643616}{\strut guy} \colorbox{red!0.9670546}{\strut know} \colorbox{blue!0.08358834}{\strut that} \colorbox{blue!3.1404097}{\strut i} \colorbox{blue!12.8370075}{\strut like} \colorbox{red!1.0698664}{\strut him} \colorbox{red!0.044517398}{\strut ?} \colorbox{blue!0.16686015}{\strut ;} \colorbox{red!0.109523594}{\strut there} \colorbox{red!0.08071572}{\strut '} \colorbox{red!0.20731191}{\strut s} \colorbox{blue!0.1495978}{\strut this} \colorbox{red!0.25173697}{\strut guy} \colorbox{blue!0.012126959}{\strut that} \colorbox{blue!0.37547377}{\strut takes} \colorbox{blue!0.12512237}{\strut a} \colorbox{red!2.0991712}{\strut class} \colorbox{red!0.004021785}{\strut with} \colorbox{blue!1.1015154}{\strut me} \colorbox{red!0.025608588}{\strut .} \colorbox{blue!0.07109939}{\strut he} \colorbox{blue!0.013496669}{\strut '} \colorbox{blue!0.12906349}{\strut s} \colorbox{blue!0.14693293}{\strut really} \colorbox{blue!2.3020601}{\strut nice} \colorbox{blue!0.28603688}{\strut and} \colorbox{blue!0.5180071}{\strut we} \colorbox{blue!0.9004262}{\strut talk} \colorbox{red!0.06481828}{\strut every} \colorbox{blue!0.43397275}{\strut day} \colorbox{blue!0.31235507}{\strut .} \colorbox{blue!0.17804812}{\strut i} \colorbox{red!0.23571087}{\strut like} \colorbox{red!97.81294}{\strut wrestling} \colorbox{blue!0.6607906}{\strut and} \colorbox{blue!0.68499947}{\strut he} \colorbox{blue!0.71631604}{\strut does} \colorbox{red!0.154433}{\strut too} \colorbox{blue!0.28372994}{\strut and} \colorbox{blue!0.09500474}{\strut we} \colorbox{blue!0.19312489}{\strut talk} \colorbox{red!0.24581671}{\strut about} \colorbox{red!0.08789354}{\strut that} \colorbox{blue!0.031725813}{\strut until} \colorbox{blue!0.011759753}{\strut the} \colorbox{red!0.609358}{\strut class} \colorbox{red!0.0038357165}{\strut starts} \colorbox{red!0.042640906}{\strut after} \colorbox{blue!0.060204662}{\strut that} \colorbox{blue!0.14232758}{\strut ,} \colorbox{red!0.004675146}{\strut i} \colorbox{red!0.06706704}{\strut don} \colorbox{red!0.1642452}{\strut '} \colorbox{red!0.0643941}{\strut t} \colorbox{red!0.08909331}{\strut see} \colorbox{red!0.09087414}{\strut him} \colorbox{red!0.24226294}{\strut anymore} \colorbox{red!0.09520022}{\strut until} \colorbox{red!0.26917547}{\strut we} \colorbox{red!0.20794016}{\strut have} \colorbox{red!0.40045962}{\strut the} \colorbox{red!1.3640548}{\strut class} \colorbox{red!0.32993644}{\strut again} \colorbox{blue!0.07871177}{\strut .} \colorbox{red!0.013045134}{\strut what} \colorbox{red!0.16953503}{\strut should} \colorbox{red!0.112107955}{\strut do} \colorbox{red!0.04067201}{\strut to} \colorbox{red!0.040416814}{\strut make} \colorbox{blue!0.3900017}{\strut him} \colorbox{blue!0.11515891}{\strut notice} \colorbox{red!0.069408715}{\strut that} \colorbox{blue!0.43267092}{\strut i} \colorbox{blue!5.3216143}{\strut like} \colorbox{red!0.29399467}{\strut him} \colorbox{red!0.027799208}{\strut ?} \colorbox{red!0.25754982}{\strut help} \colorbox{blue!0.38662016}{\strut pleasee} \colorbox{red!0.086822025}{\strut !} \colorbox{red!0.10607003}{\strut !} \colorbox{red!0.117388204}{\strut !} \colorbox{blue!0.12036127}{\strut ;} \colorbox{blue!0.074395575}{\strut well} \colorbox{red!0.47360957}{\strut u} \colorbox{red!0.2109684}{\strut should} \colorbox{red!0.12679866}{\strut try} \colorbox{red!0.07831517}{\strut to} \colorbox{red!0.29546225}{\strut stop} \colorbox{red!0.21930684}{\strut him} \colorbox{red!0.30174392}{\strut in} \colorbox{red!0.23298275}{\strut the} \colorbox{red!0.7181836}{\strut hall} \colorbox{red!0.07204509}{\strut and} \colorbox{red!0.08902114}{\strut try} \colorbox{blue!0.00856621}{\strut to} \colorbox{blue!0.07788888}{\strut say} \colorbox{blue!0.65083086}{\strut hi}  \colorbox{red!1.9188163}{\strut also} \colorbox{red!0.063688435}{\strut when} \colorbox{red!0.23232587}{\strut u} \colorbox{red!0.11759591}{\strut see} \colorbox{blue!0.015848106}{\strut him} \colorbox{red!0.1213481}{\strut try} \colorbox{red!0.032360125}{\strut to} \colorbox{blue!7.1495185}{\strut flirt} \colorbox{red!0.05898192}{\strut a} \colorbox{blue!0.12182765}{\strut little}  \colorbox{red!3.2922454}{\strut just} \colorbox{red!0.07492823}{\strut make} \colorbox{red!0.06298687}{\strut sure} \colorbox{red!0.015095965}{\strut its} \colorbox{red!0.17733103}{\strut not} \colorbox{red!0.16114454}{\strut too} \colorbox{red!0.06923199}{\strut much}  \colorbox{blue!0.066437565}{\strut a} 
}}}
\end{CJK*} & Family \& Relationships \\
\midrule
\begin{CJK*}{UTF8}{gbsn}
{\setlength{\fboxsep}{0pt}\colorbox{white!0}{\parbox{0.8\textwidth}{
\colorbox{red!0.26936495}{\strut can} \colorbox{red!1.7107216}{\strut anyone} \colorbox{red!0.57088304}{\strut tell} \colorbox{red!0.7700847}{\strut me} \colorbox{red!0.6844758}{\strut the} \colorbox{blue!0.58837616}{\strut address} \colorbox{red!0.53158635}{\strut .} \colorbox{red!0.42525196}{\strut .} \colorbox{red!0.5442199}{\strut .} \colorbox{red!0.6253268}{\strut ?} \colorbox{red!0.6807527}{\strut ;} \colorbox{red!0.8570373}{\strut to} \colorbox{blue!0.73062944}{\strut reach} \colorbox{red!2.8859365}{\strut the} \colorbox{red!8.287979}{\strut dixie} \colorbox{red!57.136154}{\strut chicks} \colorbox{red!0.48050317}{\strut by} \colorbox{red!0.5846779}{\strut ?} \colorbox{red!0.22088745}{\strut this} \colorbox{red!0.18795271}{\strut is} \colorbox{red!0.17026427}{\strut a} \colorbox{red!0.15110292}{\strut serious} \colorbox{blue!1.4226975}{\strut question} \colorbox{red!0.54728496}{\strut ,} \colorbox{red!0.90029687}{\strut so} \colorbox{red!0.55955756}{\strut please} \colorbox{red!0.57562476}{\strut don} \colorbox{red!0.12180739}{\strut '} \colorbox{red!0.1945456}{\strut t} \colorbox{red!0.6549546}{\strut post} \colorbox{blue!0.018772263}{\strut whether} \colorbox{red!0.31761566}{\strut or} \colorbox{red!0.20810407}{\strut not} \colorbox{red!1.03643}{\strut you} \colorbox{blue!2.5953498}{\strut support} \colorbox{blue!0.39951742}{\strut them} \colorbox{red!0.11019653}{\strut about} \colorbox{red!0.8352523}{\strut their} \colorbox{blue!2.8345776}{\strut comments} \colorbox{red!2.3566258}{\strut on} \colorbox{blue!65.89532}{\strut bush} \colorbox{red!0.32707593}{\strut .} \colorbox{red!0.33613226}{\strut all} \colorbox{red!0.45194203}{\strut i} \colorbox{red!0.10824522}{\strut need} \colorbox{red!0.2645694}{\strut is} \colorbox{red!0.35487854}{\strut the} \colorbox{red!0.08678499}{\strut address} \colorbox{red!0.36543643}{\strut .} \colorbox{red!0.6590609}{\strut thank} \colorbox{red!0.3675289}{\strut you} \colorbox{red!0.35426897}{\strut !} \colorbox{red!0.1664213}{\strut ;} \colorbox{red!0.2978757}{\strut hello} \colorbox{red!0.25321558}{\strut ,}   \colorbox{red!0.5126662}{\strut i} \colorbox{red!0.30631423}{\strut was} \colorbox{red!0.13634941}{\strut not} \colorbox{red!0.12848482}{\strut able} \colorbox{red!0.11619973}{\strut to} \colorbox{red!0.25004238}{\strut find} \colorbox{red!0.069323614}{\strut an} \colorbox{red!0.05478627}{\strut actual} \colorbox{blue!1.0582069}{\strut address} \colorbox{red!0.16537473}{\strut ,} \colorbox{red!0.15568666}{\strut but} \colorbox{red!0.29631868}{\strut i} \colorbox{red!0.2114558}{\strut did} \colorbox{red!0.23765643}{\strut find} \colorbox{blue!0.29177445}{\strut their} \colorbox{red!1.3892425}{\strut website} \colorbox{red!0.1788494}{\strut where} \colorbox{red!0.46462467}{\strut you} \colorbox{red!0.26177603}{\strut can} \colorbox{red!0.14061835}{\strut sign} \colorbox{red!0.2203344}{\strut up} \colorbox{red!0.14088915}{\strut for} \colorbox{blue!0.13589291}{\strut their} \colorbox{blue!1.2099824}{\strut mailing} \colorbox{red!0.38974315}{\strut list} \colorbox{red!0.2863857}{\strut and} \colorbox{red!0.19731529}{\strut i} \colorbox{red!0.24190193}{\strut did} \colorbox{red!0.18675348}{\strut find} \colorbox{red!0.21524271}{\strut this} \colorbox{blue!0.017994443}{\strut information} \colorbox{red!0.21445826}{\strut as} \colorbox{red!0.18070337}{\strut well} \colorbox{red!0.36539078}{\strut :}  \colorbox{blue!0.31916323}{\strut the} \colorbox{red!4.0166416}{\strut dixie} \colorbox{red!30.62341}{\strut chicks} \colorbox{red!0.4030672}{\strut have} \colorbox{red!0.06201823}{\strut very} \colorbox{red!0.29965585}{\strut recently} \colorbox{red!0.43527782}{\strut changed} \colorbox{red!0.82575846}{\strut management} \colorbox{red!0.6059449}{\strut .} \colorbox{red!0.07964332}{\strut i} \colorbox{red!0.1377881}{\strut do} \colorbox{red!0.053929012}{\strut not} \colorbox{red!0.1693099}{\strut yet} \colorbox{red!0.16007791}{\strut have} \colorbox{red!0.08102807}{\strut a} \colorbox{red!0.4546634}{\strut new} \colorbox{blue!0.19971752}{\strut address} \colorbox{red!0.0557623}{\strut for} \colorbox{red!24.662643}{\strut fan} \colorbox{blue!1.0502055}{\strut mail} \colorbox{red!0.1801717}{\strut .} \colorbox{red!0.25445247}{\strut once} \colorbox{red!0.7443061}{\strut one} \colorbox{red!0.1686683}{\strut is} \colorbox{red!0.50485325}{\strut available} \colorbox{red!0.08124744}{\strut ,} \colorbox{red!0.23817606}{\strut i} \colorbox{red!0.3150399}{\strut will} \colorbox{red!0.59044635}{\strut post} 
}}}
\end{CJK*} & Politics \& Government \\
\bottomrule
\end{tabular}
\end{small}
}
\caption{The top-1 memorized training instances for each class from Yahoo! Answer.
Highlighted words are those with high attribution values (red for positive memorization attribution and blue for negative memorization attribution) as computed by our memorization attribution method.
}
\label{table.vis.yahoo.attr}
\end{table*}

\end{document}